\pdfoutput=1
\documentclass[11pt]{article}
\usepackage[preprint]{acl}
\makeatletter
\@ifpackageloaded{lineno}{%
\@ifpackagelater{lineno}{2026/03/07}{}{%
  \ifdefined\IfFormatAtLeastTF
  \IfFormatAtLeastTF{2025-06-01}{%
    \RemoveFromHook{build/column/before}[lineno]
    \AddToHook{build/column/after}[acl-lineno-fix]{%
      \setbox\@outputbox\vbox to\ht\@outputbox{%
        \boxmaxdepth\@maxdepth
        \protected@write\@auxout{}{%
          \string\def\string\@LN@column{\if@firstcolumn1\else2\fi}%
        }%
        \unvbox\@outputbox
      }%
    }%
  }{}%
  \fi
}
}{}
\makeatother
\usepackage{times}
\usepackage{latexsym}
\usepackage[T1]{fontenc}
\usepackage[utf8]{inputenc}
\usepackage{microtype}
\IfFileExists{inconsolata.sty}{\usepackage{inconsolata}}{}  %
\usepackage{graphicx}
\usepackage{booktabs}
\usepackage{amsmath,amssymb}
\usepackage{xcolor}
\usepackage{colortbl}
\usepackage{multirow}
\usepackage{enumitem}
\usepackage{placeins}

\newif\ifprovmark \provmarkfalse
\newcommand{\TODO}[1]{\textcolor{red}{[TODO: #1]}}
\newcommand{\method}{\mbox{FlagRot}}
\newcommand{\mkivi}{\mbox{FlagRot-KIVI}}
\newcommand{\mauto}{\mbox{FlagRot-auto}}

\newcommand{\mlarge}{\method{}-L}
\newcommand{\figpdf}[2]{\IfFileExists{#2}{\includegraphics[width=#1]{#2}}{\fbox{\parbox[c][4cm][c]{\dimexpr#1-2\fboxsep-2\fboxrule\relax}{\centering\TODO{figure} \texttt{\detokenize{#2}}}}}}
\newcommand{\turboBitsRange}{4.9--7.1}
\newcommand{\sotaGmCalib}{7.02}
\newcommand{\sotaGmOurs}{7.88}
\newcommand{\kvFourBitRangeAll}{13--89}
\newcommand{\kvLargeRangeAll}{34--94}
\newcommand{\nModels}{15}
\newcommand{\oursVsTurbo}{29--73}
\newcommand{\oursVsSmooth}{7--57}
\newcommand{\nvPlusOurs}{15--55}
\newcommand{\nvOursVsRtn}{17--56}
\newcommand{\dynWins}{40 of 45}
\newcommand{\kvMxQwenFull}{41}
\newcommand{\kvNvQwenFull}{215}
\newcommand{\valueOnly}{4--5}
\newcommand{\dseTwoKeys}{56}
\newcommand{\dseBlockHead}{59}
\newcommand{\keyShare}{93--94}
\newcommand{\dseStaticKl}{43}
\newcommand{\bokTwo}{0.64}
\newcommand{\bokSixtyFour}{0.30}
\newcommand{\dseCbInt}{73}
\newcommand{\bitConsistency}{0.71--0.82}
\newcommand{\mechN}{15}
\newcommand{\wsC}{19--81}
\newcommand{\wsCK}{14--81}
\newcommand{\wsWushGm}{2.33}
\newcommand{\wsCGm}{1.51}
\newcommand{\wsN}{12}
\newcommand{\baseCOverW}{3--15}
\newcommand{\qcRtnPlain}{4697}
\newcommand{\qcOursVsRtn}{43--76}
\newcommand{\qcOursVsRtnSw}{30--43}
\newcommand{\kvMedInt}{56}
\newcommand{\kvMedMx}{67}
\newcommand{\kvMedNv}{44}
\newcommand{\swOursNone}{7.88}
\newcommand{\swOursWin}{2.08}
\newcommand{\swOursDef}{1.48}
\newcommand{\swOursVsRtn}{47}
\newcommand{\swOursVsStatic}{21}
\newcommand{\swSixteenGain}{1}
\newcommand{\swFp}{3.71}
\newcommand{\lbvOursLlama}{$-$0.12}
\newcommand{\lbvHtoh}{+0.49 [+0.01, +0.96]}
\newcommand{\lbvOursQwen}{$-$0.01}
\newcommand{\lbvHtohQwen}{$-$0.25 [$-$0.62, +0.13]}
\newcommand{\lbvOursBig}{$-$0.16}
\newcommand{\lbvHtohBig}{$-$0.18 [$-$0.40, +0.04]}
\newcommand{\tpOursFiH}{0.92--0.94}
\newcommand{\tpOursFiL}{0.83--0.96}
\newcommand{\tpStaticFiH}{0.89--1.00}
\newcommand{\tpStaticFiL}{0.81--0.90}
\newcommand{\tpOursWzH}{0.84--0.92}
\newcommand{\tpOursWzL}{0.95--1.00}
\newcommand{\tpOursBfH}{3.2}
\newcommand{\tpOursBfL}{6.2}
\newcommand{\tpOursBfLmid}{4.5}
\newcommand{\tpOursRtn}{0.91--1.01}
\newcommand{\motFlatUp}{65--72}
\newcommand{\motSpikyUp}{14--19}
\newcommand{\motNvUp}{38--60}
\newcommand{\motShareHead}{40--43}
\newcommand{\motShareBlock}{37--39}
\newcommand{\motShareId}{18--23}
\newcommand{\ctOurs}{56}
\newcommand{\ctOursSw}{46}
\newcommand{\ctOursNv}{46}
\newcommand{\ctOursNvSw}{46}
\newcommand{\ctMse}{41}
\newcommand{\ctMseSw}{35}
\newcommand{\ctMseNv}{23}
\newcommand{\ctMseNvSw}{27}
\newcommand{\ctStatic}{41}
\newcommand{\mechFpQwen}{+264}
\newcommand{\mechUnifQwen}{$-$27}
\newcommand{\mechSign}{14}
\newcommand{\mechRho}{0.94}
\newcommand{\mechRatioBefore}{0.59--0.99}
\newcommand{\mechRatioAfter}{1.21--2.55}
\newcommand{\tkBOursN}{36}
\newcommand{\tkBOursIn}{34}
\newcommand{\tkBOursHi}{2}
\newcommand{\iEoneSel}{1.4}
\newcommand{\iLbRtnQwenAbs}{11.0}
\newcommand{\iRulerSelD}{4.2}
\newcommand{\iQFaTwo}{0.88}
\newcommand{\iQFpTwo}{1.03}
\newcommand{\iQBfPre}{566/641}
\newcommand{\iQPage}{784/1568/5024}
\newcommand{\iVklKivi}{18.0}
\newcommand{\iVklStatic}{14.7}
\newcommand{\iVklDyn}{12.3}
\newcommand{\iKlKivi}{18.4}
\newcommand{\iKlFrk}{12.3}
\newcommand{\iKlStaticWorse}{11}
\newcommand{\iAimeSelAutoD}{+14.6}
\newcommand{\iLcQwenFaGrow}{3.5}
\newcommand{\iLcQwenFrkGrow}{1.4}
\newcommand{\iWsGrowQwen}{1.6}
\newcommand{\iSoftKivi}{+5}
\newcommand{\iSoftKiviSel}{+10}
\newcommand{\iKiviCenterRange}{$-$0.6 to +6.4}
\newcommand{\iCrFc}{21.9}
\newcommand{\iCrFcRot}{23.0}
\newcommand{\iCrKc}{11.7}
\newcommand{\iCrKcRot}{12.0}
\newcommand{\iTqThree}{11.4 to 10.1}
\newcommand{\iTqFour}{9.2 to 7.9}
\newcommand{\iHfVlQwen}{+0.11}
\newcommand{\iHfVlLlama}{+0.40}
\newcommand{\iFamHfFixD}{+2.58}
\newcommand{\iPkFrkTw}{12.4}
\newcommand{\iPkFrkVsRkTwB}{9}
\newcommand{\iPkKiviVsFaB}{2}
\newcommand{\iHfRkOneQwen}{$-$2.17 [$-$3.43, $-$0.94]}
\newcommand{\iHfRkFaithQwen}{$-$12.67}
\newcommand{\iHfRkOneLlama}{$-$2.40 [$-$3.77, $-$1.09]}
\newcommand{\iHfRkFaithLlama}{$-$7.26}
\newcommand{\iRdTwFrVsRtn}{55--81}
\newcommand{\iRdTwSelVsKivi}{24--54}
\newcommand{\iRdTwSelLowest}{each}
\newcommand{\iDomFr}{28--90}
\newcommand{\iDomFrk}{77--97}
\newcommand{\iDomFrkBelowKivi}{each}
\newcommand{\iCostDecPct}{4.8}
\newcommand{\iCostDecPctFour}{5.3}
\newcommand{\iCostCapX}{3.0}
\newcommand{\iCostCapBf}{0.93}
\newcommand{\iCostCapTwo}{2.75}
\newcommand{\iCostPkDec}{33}
\newcommand{\iCostPkPageX}{6.3}
\newcommand{\iCostPkPrefillSel}{5}
\newcommand{\iCostPkCapX}{3.0}
\newcommand{\iFxKlStatic}{13.7}
\newcommand{\iFxKlStaticPk}{13.8}
\newcommand{\iFxRedTok}{35}
\newcommand{\iFxSharePk}{78}
\newcommand{\iFxDynBetterPk}{11}
\newcommand{\iFxRedKiviPk}{10}
\newcommand{\iDomRelMin}{24}
\newcommand{\iDomRelMax}{44}
\newcommand{\iCapPkEight}{6.1}
\newcommand{\iKivDecFast}{2.7--2.9}
\newcommand{\iRkNoGemFrk}{11.1}
\newcommand{\iRkNoGemRk}{14.1}
\newcommand{\iUFkpTwo}{12.4}
\newcommand{\iUFkpTwoHo}{12.1}
\newcommand{\iUKpTwo}{18.7}
\newcommand{\iURelFaFour}{36}
\newcommand{\iURelFaThree}{51}
\newcommand{\iURelFaTwo}{59}
\newcommand{\iURelFkpFour}{30}
\newcommand{\iURelFkpThree}{31}
\newcommand{\iURelFkpTwo}{34}
\newcommand{\iGgCrit}{29}
\newcommand{\iGgGm}{11.9}
\newcommand{\iGgGmSel}{12.3}
\newcommand{\iGgKl}{22.7}
\newcommand{\iGgLocal}{52}
\newcommand{\iGgMeas}{29}
\newcommand{\iGgOff}{27}
\newcommand{\iGgOracle}{18.3}
\newcommand{\iGgOtherMax}{2}
\newcommand{\iGgVsKivi}{9.5}
\newcommand{\iGgZero}{5}
\newcommand{\iGmBestRel}{+21}
\newcommand{\iGmBestRelLlama}{$-$0.9}
\newcommand{\iGmBestRelQwen}{$-$1.5}
\newcommand{\iGmKivi}{25.1}
\newcommand{\iPsFpWorst}{0.75}
\newcommand{\iPsRulerFa}{+18.5}
\newcommand{\iMeAgreeKivi}{62--69}
\newcommand{\iMeAgreeTok}{39--44}
\newcommand{\iMeCross}{2.27--2.35}
\newcommand{\iMeKappaAfter}{2.2}
\newcommand{\iMeKivMse}{19.1}
\newcommand{\iMeKivNone}{18.4}
\newcommand{\iCsKiviMax}{2.1}
\newcommand{\iCsFaOne}{5.8}
\newcommand{\iCsFaRest}{1.9}
\newcommand{\iTfRulerSel}{+2.9}
\newcommand{\iTfRulerFx}{+2.5}
\newcommand{\iTfLbQSel}{+0.42}
\newcommand{\iTfLbLSel}{+0.86}
\newcommand{\iTfLbQFx}{+0.72}
\newcommand{\iTfLbLFx}{+0.47}
\newcommand{\iTfThirtyTwoSel}{+1.94}
\newcommand{\iTfThirtyTwoFr}{$-$0.23}
\newcommand{\iTfThirtyTwoKiviLoss}{2.17}
\newcommand{\iTfTwentySevenSel}{$-$0.57}
\newcommand{\iTfLbLossN}{three}
\newcommand{\iTfLbN}{four}
\newcommand{\iTfLbLossRange}{+0.4 to +1.9}
\newcommand{\iTfRulerRec}{26}
\newcommand{\iTfTwentySevenFrAbs}{0.24}
\newcommand{\iCvEightFp}{1.4}
\newcommand{\iCvEightBf}{3.0}

\newcommand{\had}{\mathbf{H}}
\newcommand{\R}{\mathbf{R}}

\title{FlagRot: Selective Rotation for Low-Bit KV Caches}

\author{
Kunming Shao$^{1}$ \quad Xijie Huang$^{2\dagger*}$ \quad Jierun Chen$^{1\dagger}$ \quad Jiaqi Shao$^{1}$ \quad Yanli Wang$^{3}$ \\
\textbf{Jiangnan Yu}$^{1}$ \quad \textbf{Qiwei Li}$^{4}$ \quad \textbf{Wei Zhang}$^{1}$ \quad \textbf{Kwang-Ting Cheng}$^{1}$ \quad \textbf{Chi Ying Tsui}$^{1}$ \\[4pt]
$^{1}$The Hong Kong University of Science and Technology \quad $^{2}$Meta \\
$^{3}$Sun Yat-sen University \quad $^{4}$Wuhan University \\[2pt]
\texttt{kshaoaa@connect.ust.hk} \quad $^{\dagger}$Corresponding authors
}

\begin{document}
\maketitle
{\renewcommand{\thefootnote}{*}\footnotetext{All the work was done within HKUST and Xijie Huang served an advisory role.}}

\begin{abstract}
Long-context inference is bound by the KV cache, which 2-bit codes could shrink about sixfold if their
error stays in check. The usual remedy rotates keys before quantizing, assuming rotation always helps.
However, we observe that it does not: rotation flattens a spiky piece of a key but can give a flat one a new peak, and
both coexist within one key. We therefore propose \method{}, which stores each 32-channel piece rotated or
as is, whichever has the smaller calibrated attention error, and rotates the query instead when reading.
We further show that how often to decide follows the quantization axis: per-channel caches such as KIVI need one
decision per head, at no write cost, while per-token caches decide per key, with the choice stored in existing sign bits.
At 2.5 bits per value, \method{} lowers the twelve-model KL divergence of KIVI from \iUKpTwo{} to
\iUFkpTwo{}$\times10^{-3}$, below RotateKV on \iPkFrkVsRkTwB{} of 12 models. Served as a vLLM plugin, it raises RULER at
128k tokens by \iTfRulerSel{} points and the LongBench~v1 mean by \iTfLbLossRange{} points on the \iTfLbLossN{} of
\iTfLbN{} models where 2-bit KIVI loses accuracy. It also lowers the error of 4-bit integer and FP4 caches.\footnote{\scriptsize Code: \href{https://github.com/KunmingSHAO/FlagRot_releaser}{\mbox{github.com/KunmingSHAO/FlagRot\_releaser}}}
\end{abstract}

\section{Introduction}
\label{sec:intro}
\begin{figure*}[t]
\centering
\includegraphics[width=\textwidth]{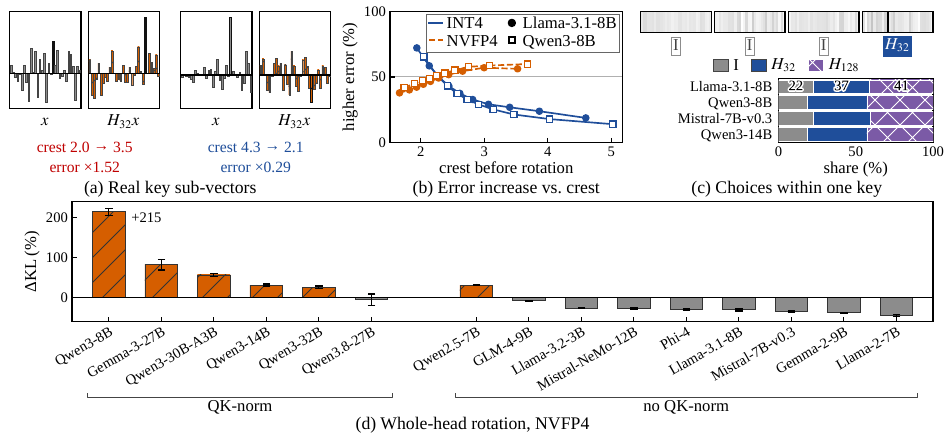}
\caption{On real cached keys, rotation helps spiky sub-vectors and hurts many flat ones. (a)~Two 32-dim key sub-vectors of Llama-3.1-8B before and after $\had_{32}$,
with crest factor and change of the query-weighted INT4 error. (b)~Share of sub-vectors whose error rises under
$\had_{32}$, by decile of the crest factor. (c)~The selections of \method{} within one key and their shares. (d)~Whole-head
rotation of NVFP4 caches, per model.}
\label{fig:motivation}
\end{figure*}
Every decoding step of a long-context model reads the whole KV cache, so the bytes per cached value bound both the
number of sequences that fit in memory and the time of each step. At 32,768 tokens the BF16 cache of Qwen3-8B takes
4.8\,GB per sequence. A cache with 2-bit codes and their scales needs 2.5 to 3 bits per value instead of 16. The price is accuracy: with
2-bit round-to-nearest (RTN) codes, the LongBench~v1 mean of Qwen3-8B falls by \iLbRtnQwenAbs{} points in our served
2.5-bit per-token setting (Table~\ref{tab:tasks2full}). Range is one source of this error: a few large entries widen the range of their quantization group and
leave few code levels for the rest \citep{liu2024kivi,hooper2024kvquant}.

Rotations attack the range problem by spreading large entries over many coordinates
\citep{ashkboos2024quarot,rotatekv2025,zandieh2026turboquant}, but we find that they spread small entries as well. The error of a
uniform quantizer grows with the crest factor of a group, and a Hadamard transform pulls every group toward the crest
factor of a Gaussian vector: it flattens a spike but can give a flat piece a peak. On the cached keys of Llama-3.1-8B, we
measure that it makes most of the flattest pieces harder to quantize (Figure~\ref{fig:motivation}). Which transform is better
changes from one piece to the next, from token to token and from model to model, so a single rotation for all keys
leaves error behind even when it helps on average.

We therefore treat the transform as a decision of the encoder rather than a fixed preprocessing step. Our method,
\method{}, tries
both versions of each 32-channel piece, plain and rotated, and selects by a calibration estimate of attention error:
a key error $e$ shifts the logit of a query $q$ by $q\cdot e$, and queries
concentrate on a few channels, so the decision weighs errors by calibration queries. One bit per piece records it in
the sign bit of a positive quantization scale in the packed layout, so its page size does not grow; attention rotates the query
into both bases instead of rotating keys back.

Selection applies to several quantizers, and we find that how often to decide depends on the quantization axis. The encoder can
decide once per head from calibration data (\emph{fixed mode}) or anew for every key (\emph{per-key mode}). Our main
system, \mkivi{}, adds selection to KIVI, which quantizes keys along channels; \emph{KIVI-only} is the same cache with
selection off. There one decision already covers 32 tokens, and the fixed mode keeps \iFxSharePk\% of the KL reduction with no
work at write time. A second instance, \mauto{}, adds selection to a cache that quantizes each key on its own after
\emph{centering}, that is, subtracting a calibrated key mean. On such per-token caches the per-key mode lowers the
error by a further \iFxRedTok\%, against \iFxRedKiviPk\% on KIVI. One principle covers both: the transform is a decision of the
encoder, judged by the error attention sees and renewed only as often as the statistics the quantizer pools change.

This paper makes three contributions. First, we explain when a rotation helps a low-bit key, through the crest
factor of each piece and the queries that read it (Section~\ref{sec:background}). Second, we introduce \method{}, which
selects the transform of each piece by the error attention sees, records the decision in existing sign bits, and runs
in vLLM as a per-channel and a per-token cache (Section~\ref{sec:method}). Third, we show that selection lowers the
error of the encoders we test: at 2.5 bits per value it lowers the twelve-model KL of KIVI from \iUKpTwo{} to
\iUFkpTwo{}$\times10^{-3}$, below RotateKV (clip 1.0) on \iPkFrkVsRkTwB{} of the models (Table~\ref{tab:unified}).
Served at that size, it adds \iTfRulerSel{} points to the RULER score at 128k tokens and \iTfLbLossRange{} points to
the LongBench~v1 mean on the \iTfLbLossN{} of \iTfLbN{} models where KIVI loses accuracy (Table~\ref{tab:tasks25}), and \mkivi{} outscores the per-token cache of equal
size on the three shared endpoints. At 4 bits our selection lowers the KL of INT4, MXFP4 and NVFP4 caches by
\kvFourBitRangeAll\% relative to RTN.

\section{When Does Key Rotation Help?}
\label{sec:background}
\begin{figure*}[t]
\centering
\figpdf{\textwidth}{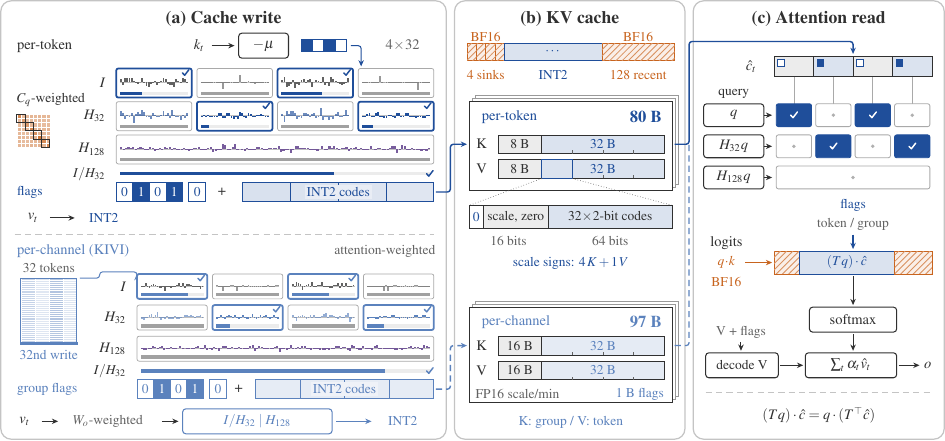}
\caption{\method{}. (a)~At cache write, each 32-dim key sub-vector keeps $I$ or $\had_{32}$, or the whole head takes
$\had_{128}$, whichever has the lower query-weighted error; the per-token branch first subtracts the calibrated key mean,
and the per-channel branch decides once per completed 32-token key group. (b)~INT2 pages of 80 and 97 bytes per token and
KV head ($D{=}128$), with the flags in the sign bits of the scales and 4 sink and 128 recent tokens in BF16. (c)~Attention
rotates the query into each basis once per step; the flags select the partial dot products, and keys are never rotated
back.}
\label{fig:overview}
\end{figure*}

Whether a rotation helps a low-bit key depends on the shape of each piece and on where the queries look. A quantizer
shares one scale, and for asymmetric formats one offset, among $g$ consecutive values. With $m$ metadata
bits per group, a $b$-bit code costs $b+m/g$ bits per value; an INT2 code with a 16-bit scale word and $g{=}32$ costs
2.5 bits. Vectors are columns. An orthogonal matrix $\R$ preserves dot products, $q^\top k=(\R q)^\top(\R k)$, so a key may be
stored in a rotated basis if the query is rotated alike. The normalized Walsh--Hadamard matrix $\had_r$ is the usual
transform: it costs $O(r\log r)$ operations and spreads every coordinate over all $r$ outputs.

Spreading helps spiky pieces and can hurt flat ones. A uniform $b$-bit quantizer divides the range of a group into
$2^b{-}1$ steps, so, as a rule of thumb, at a fixed energy its error grows with the crest factor
$\kappa(x)=\|x\|_\infty/\mathrm{rms}(x)$. An orthogonal $\had_r$ keeps the energy and replaces $\kappa(x)$ by
$\kappa(\had_r x)$, and on the keys we measure, rotated pieces settle at a median crest factor of \iMeKappaAfter, near
the 2.3 of a Gaussian vector of 32 entries. Rotation thus pulls every piece toward that value: a spike ($\kappa{=}\sqrt r$) becomes flat, but a flat piece ($\kappa$ near 1) can gain a peak
(Figure~\ref{fig:motivation}a). Among the flattest tenth of the key pieces of Llama-3.1-8B and Qwen3-8B, rotation makes
the error worse for about two thirds; among the spikiest tenth it does so for about one in six
(Figure~\ref{fig:motivation}b). With 2-bit codes, the share of pieces that rotation hurts crosses one half at
$\kappa$ of \iMeCross{} on the four development models. The crest factor captures the trend; the exact error also depends on the min--max range and, in FP4, on a
non-uniform grid, so \method{} quantizes each candidate and measures it. Models that normalize their keys with QK-norm have flat keys throughout, and rotating
whole heads hurts most of them (Figure~\ref{fig:motivation}d). Within a single key the pieces disagree
(Figure~\ref{fig:motivation}c), and the same piece of the same head prefers different transforms on different tokens.

Reconstruction error is also the wrong yardstick. A key error $e$ changes the logit of a query $q$ by $q^\top e$, so what
attention sees is $\mathbb{E}_q[(q^\top e)^2]=e^\top C_q e$ with $C_q=\mathbb{E}[q q^\top]$, and the queries of a head
concentrate their energy on a few channels. A rotation that shrinks the error of a piece spreads what remains evenly
over its channels, which can move error into exactly the channels the queries read. The decision should thus weigh
errors by the queries, and how often it must be renewed depends on how much the statistics behind it change from one
key to the next.

Keys have one more property that matters for 2-bit codes: after RoPE, the keys of many heads share a large common
offset. Subtracting the same vector $\mu$ from every key changes every attention logit of a query by the same amount,
$q\cdot\mu$, so the attention weights do not change (for models without logit soft-capping). A quantizer that works on
one key at a time wastes part of its range on this offset; KIVI, which works along channels, absorbs it.

\section{\method{}}
\label{sec:method}

\subsection{Selecting the Transform}
\label{sec:select}
We treat the transform as part of the code. For a vector or key group $x$ and a candidate set $\mathcal T$, the
encoder picks
\begin{gather}
T^\ast(x)=\arg\min_{T\in\mathcal T}\, e_T^\top M e_T,\label{eq:select}\\
e_T=x-T^{-1}Q(Tx),\nonumber
\end{gather}
where $Q$ is the quantizer and $M$ a positive semi-definite weight derived from the queries (Section~\ref{sec:instances}).
A head of dimension $D$ is split into pieces of $r{=}32$ channels; each piece may stay as it is or take $\had_r$, and the
whole head may instead take $\had_D$. Each piece $j$ keeps the candidate with the lower $e_j^\top M_{jj} e_j$, and the
assembled reconstruction then competes with the $\had_D$ reconstruction under the full $e^\top M e$, giving $D/r+1$
flags, five for $D{=}128$. The piece-wise choice is close to the best joint one: searching all 16 combinations of the
four pieces changes the KL of the two 8B models by \iGmBestRelLlama\% and \iGmBestRelQwen\% (Appendix~\ref{app:int2-sim}). Read as a coding problem, Eq.~\ref{eq:select} trades side information for distortion: a
per-key decision costs $D/r+1$ bits per key, a fixed one none.

At read time the inverse transform moves to the query:
\begin{equation}
q\cdot\hat k_t=\sum_{j}\bigl(T_{t,j}\,q_j\bigr)\cdot\hat c_{t,j},
\label{eq:attn}
\end{equation}
where $\hat c_{t,j}$ is the dequantized code of piece $j$ and $T_{t,j}\in\{I,\had_r\}$ is its stored transform; with
the head flag set, the product is $(\had_Dq)\cdot\hat c_t$. The query is transformed into both bases once per decoding
step and shared by all cached tokens. Value transforms are applied after the attention-weighted sum, separately per
basis. The weight $M$ and all other statistics come from four calibration windows and are frozen before evaluation, so
a decision depends only on the tokens it encodes. Codes and flags are written once and never revised.

The same criterion supports two modes. In the \emph{fixed} mode, the decision for each layer, KV head and piece is
taken once on the calibration windows and stays the same for every token; the flags are then constants and writing a
key takes a single quantization. In the \emph{per-key} mode, the encoder decides anew for every key it
writes; it quantizes each key or key group three times (plain, with $\had_r$ per piece and with $\had_D$) and keeps the flags of
the winner. Both instances below use the per-key mode unless stated otherwise.

\subsection{FlagRot-KIVI and a Per-Token Instance}
\label{sec:instances}
The quantization axis determines what one decision covers. Our main system, \mkivi{}, follows KIVI and quantizes keys per channel over groups of 32 tokens, with an FP16 scale and minimum per channel
and group \citep{liu2024kivi}. A group is encoded when its last token is written; until then its keys stay in the BF16
window. One decision covers a whole group of one KV head, and its error is weighted by how strongly the calibration
queries attend to each key:
\begin{gather}
\varepsilon_G=\sum_{t\in G}\sum_{q} w_{t,q}\,(q\cdot e_t)^2,\\
w_{t,q}=\min\bigl(1,\exp(s\,q\cdot k_t-\log Z_q)\bigr),
\end{gather}
where $s$ is the logit scale and $Z_q$ the softmax normalizer of $q$ over its calibration window. The weight is the
attention probability the calibration query would assign to $k_t$; it is capped at one because $Z_q$ is measured on a
different window and can be smaller than the true normalizer. Values are quantized per token, and their decisions weight
the error by the same attention mass and by the column norms of the output projection. Because per-channel offsets absorb
a shared key mean, \mkivi{} needs no centering.

The second instance, \mauto{}, quantizes each key per token in 32-channel groups and weights the error by the second moment of the
calibration queries, $M=C_q=\mathbb{E}[q q^\top]$. Before encoding it removes the shared key offset of Section~\ref{sec:background}: it subtracts $\mu\odot m$,
where $\mu$ is the calibrated post-RoPE key mean of the head and the mask $m\in\{0,1\}^D$ keeps the 32-channel groups in
which removing the mean lowers the calibration error of an RTN cache. The BF16 sink and recent keys receive the same
shift. If removing the mean raises the calibration KL of a model, the model keeps its keys unshifted; this automatic
switch is the \emph{auto} in the name. Values are
quantized per token without rotation.

With $D{=}128$, a token of one KV head occupies 512 bytes in BF16. \mkivi{} with FP16 scales and minima, which cost
32 bits per group, needs 97 bytes, the \emph{3.0-bit layout}. Storing them in FP8 gives the \emph{packed} 2.5-bit
layout of 80 bytes, the size of an RTN page and of \mauto{}; we serve both. Both instances run in vLLM
\citep{kwon2023vllm} with paged INT2 storage, a BF16 buffer for 4 sink and 128 recent tokens (S4W128), CUDA graphs and
prefix caching.

\section{Experimental Setup}
\label{sec:setup}
We measure fidelity as the next-token KL divergence from the BF16 model, reported $\times10^3$ and averaged over all
2048-token WikiText-2 test windows, on twelve decoder-only models from 3B to 27B parameters
(Appendix~\ref{app:int2}). Keys after RoPE and values are quantized, linear layers stay in BF16, and every unprotected
earlier token is read from the quantized cache. Calibration uses four 2048-token windows of the WikiText-2 training
split \citep{merity2017wikitext}. We developed the method on Llama-3.1-8B, Qwen3-8B, Mistral-7B-v0.3 and Qwen3-14B; the
other eight models are held out and were run once with the final settings. Unless stated
otherwise, every INT2 cache uses $g{=}32$ and S4W128.

Task evaluations run our kernels in vLLM~0.30. LongBench~v1 with all 21 tasks and official prompts
\citep{bai2024longbench} covers four models: Qwen3-8B, Llama-3.1-8B-Instruct, Qwen3.8-27B \citep{qwen2026qwen38} and
the dense Qwen3-32B. RULER with 13 subtasks at 128k tokens \citep{hsieh2024ruler} runs on Llama-3.1-8B-Instruct, the
8B model with a native 128k context. Reasoning with thinking (AIME 2024 and 2025 on Qwen3-8B) enters the per-token
study of \S\ref{sec:attr}, agentic tasks ($\tau^2$-bench \citep{barres2025tau2}, SWE-bench Verified) the 4-bit study
of \S\ref{sec:four}, and both are reported in full in the appendix. A cache and its BF16
reference always run on the same GPU type.

KIVI-only fixes every flag of \mkivi{} to the identity and shares its kernel, layout and protection; the fixed mode
uses the same candidates and criterion as the per-key mode. RotateKV \citep{rotatekv2025} is our simulation, validated against the perplexity of its official
code, and runs with our protection at clip ratios 0.8 and 1.0. WUSH-KV runs in its published configuration
\citep{wushkv2026}. We report nominal 95\% paired bootstrap intervals over windows, items within tasks, problems or
tasks. We fixed two families of task contrasts before the last task results were in; within them, significance means a
Holm-adjusted $p<0.05$, and other intervals are descriptive.

\section{Results}
\label{sec:results}

\subsection{Fidelity Across Encoders and Bit Rates}
\label{sec:fid}
\begin{table}[t]
\centering
\small
\setlength{\tabcolsep}{1pt}
\begin{tabular*}{\columnwidth}{@{\extracolsep{\fill}}lrrr@{}}
\toprule
\multicolumn{4}{@{}l}{(a) Equal size: INT2 at 2.5 bits per value} \\
\midrule
Cache & KL & Held-out & \mkivi{} L/H \\
\midrule
KIVI & 18.7 & 17.6 & 11/1 \\
RotateKV, clip 0.8 & 25.6 & 26.5 & 12/0 \\
RotateKV, clip 1.0 & 18.1 & 19.5 & 9/2 \\
\mauto{} (ours) & 17.8 & 17.8 & 12/0 \\
\mkivi{} (ours) & \textbf{12.4} & \textbf{12.1} & -- \\
\bottomrule
\end{tabular*}
\par\medskip
\begin{tabular*}{\columnwidth}{@{\extracolsep{\fill}}lrrr@{}}
\toprule
\multicolumn{4}{@{}l}{(b) Selection on each encoder, by code width} \\
\midrule
Encoder & INT2 & INT3 & INT4 \\
\midrule
RTN & 75.4 & 10.5 & 2.8 \\
\quad + sel. & \textbf{28.5} {\scriptsize $-$62\%} & \textbf{4.6} {\scriptsize $-$56\%} & \textbf{1.5} {\scriptsize $-$47\%} \\
RTN, centered & 43.4 & 6.7 & 2.0 \\
\quad + sel. & \textbf{17.8} {\scriptsize $-$59\%} & \textbf{3.3} {\scriptsize $-$51\%} & \textbf{1.3} {\scriptsize $-$36\%} \\
KIVI, FP8 & 18.7 & 3.7 & 1.7 \\
\quad + sel. & \textbf{12.4} {\scriptsize $-$34\%} & \textbf{2.5} {\scriptsize $-$31\%} & \textbf{1.2} {\scriptsize $-$30\%} \\
KIVI, FP16$^\dagger$ & 18.4 & 3.2 & 1.2 \\
\quad + sel. & \textbf{12.3} {\scriptsize $-$33\%} & \textbf{2.4} {\scriptsize $-$24\%} & \textbf{1.0} {\scriptsize $-$14\%} \\
\bottomrule
\end{tabular*}
\caption{Simulated KL$\times10^3$ ($\downarrow$) with groups of 32 and S4W128, geometric mean over twelve models and over the eight held-out ones. (a)~Caches of equal size; L/H: models on which \mkivi{} is significantly lower / higher than the row (paired 95\% CI). (b)~Each encoder without and with selection (+ sel.); percentages relative to the row above, at $b{+}0.5$ bits per value with FP8 or FP16 scale and minimum ($^\dagger b{+}1$). Models significantly lower with selection: 10--12 of 12 per cell.}
\label{tab:unified}
\end{table}

\begin{figure*}[t]
\centering
\figpdf{0.36\textwidth}{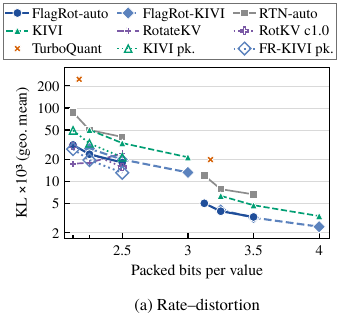}\hfill\figpdf{0.62\textwidth}{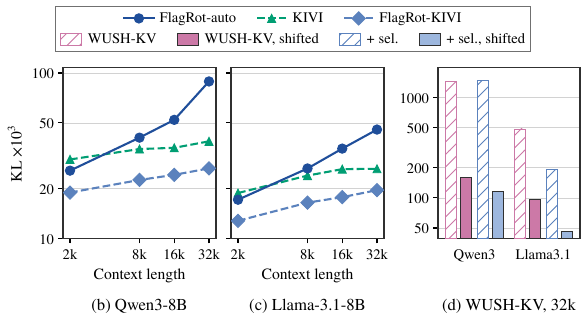}
\caption{Simulated KL$\times10^3$ with S4W128. (a)~Geometric mean over the four development models against nominal
bits per value; RTN-auto: RTN with the centering of \mauto{}. (b,~c)~Context length; 8k, 16k and 32k use 8, 6 and 4 windows. (d)~WUSH-KV's
fitted transform at 32k, without and with selection and RoPE-shifted calibration.}
\label{fig:rdlen}
\end{figure*}
At equal size, our \mkivi{} has the lowest error of the 2-bit caches we compare (Table~\ref{tab:unified}a). In the
packed layout it fits the 2.5 bits of a per-token page and reaches a KL of \iPkFrkTw, below KIVI on 11 of the twelve
models, below \mauto{} on 12 and below RotateKV with clip 1.0, the stronger of its two settings, on
\iPkFrkVsRkTwB{} of them. Without Gemma-3-27B, where RotateKV degrades, the geometric means are
\iRkNoGemFrk{} and \iRkNoGemRk. On the eight held-out models, \mkivi{} reaches a geometric mean of \iUFkpTwoHo.

Selection lowers the KL of the four encoders at INT2, INT3 and INT4, by more at lower widths
(Table~\ref{tab:unified}b). On KIVI it removes \iURelFkpTwo\% of the KL at INT2, and \iURelFkpThree\% and
\iURelFkpFour\% at INT3 and INT4; on centered per-token caches it removes \iURelFaTwo, \iURelFaThree{} and
\iURelFaFour\%. Without selection, packed KIVI beats \mauto{} on \iPkKiviVsFaB{} models. The one model on which
selection raises the INT2 KL of KIVI, Gemma-3-27B, can be told from the calibration windows alone; a per-layer check
there brings it to \iGgVsKivi\% below KIVI and keeps the other models within their paired intervals (Appendix~\ref{app:int2-sim}).
Selection also composes with fitted transforms: started from the transform fitted by WUSH-KV, or from RotateKV under
our protection, it lowers the KL of the models we ran (Appendix~\ref{app:int2-sim}).

The per-channel layout also ages better with context. From 2k to 32k tokens on Qwen3-8B, the KL of \mkivi{} grows by
\iLcQwenFrkGrow$\times$ and \mauto{} grows by \iLcQwenFaGrow$\times$ (Figure~\ref{fig:rdlen}b).

Calibration on WikiText-2 transfers to other text: with these statistics, selection lowers the KL of KIVI
by \iDomRelMin--\iDomRelMax\% on books, code, math, Chinese and German Wikipedia and 32k-token windows
(Appendix~\ref{app:int2-sim}); other calibration windows, sources or query counts change it by at most \iCsKiviMax\%
(Table~\ref{tab:calibsens}).

\subsection{Task Accuracy in vLLM}
\label{sec:tasks}
\begin{table*}[t]
\centering
\small
\setlength{\tabcolsep}{3pt}
\renewcommand{\arraystretch}{1.15}
\begin{tabular}{@{}p{\dimexpr0.145\textwidth-3pt\relax}p{\dimexpr0.19\textwidth-6pt\relax}>{\raggedleft\arraybackslash}p{\dimexpr0.225\textwidth-6pt\relax}>{\raggedright\arraybackslash}p{\dimexpr0.195\textwidth-6pt\relax}>{\raggedright\arraybackslash}p{\dimexpr0.245\textwidth-3pt\relax}@{}}
\toprule
Benchmark & Model & {\footnotesize BF16 $\to$ KIVI $\to$ Ours} & \multicolumn{1}{c}{Fixed-mode gain} & \multicolumn{1}{c}{\textbf{Per-key gain}} \\
\midrule
RULER 128k & Llama-3.1-8B-It & {\footnotesize 73.7$\,\to\,$62.3$\,\to\,$65.2} & \mbox{\makebox[2.8em][r]{+2.5}\hspace{0.3em}{\scriptsize\textcolor{black!65}{[+0.9, +4.1]}}} & \mbox{\makebox[2.8em][r]{\textcolor{darkblue}{\textbf{+2.9}}}\hspace{0.3em}{\scriptsize\textcolor{black!65}{[+1.2, +4.5]}}} \\
\addlinespace[4pt]
\multirow[t]{4}{*}{LongBench v1} & Qwen3-8B & {\footnotesize 49.35$\,\to\,$47.28$\,\to\,$47.70} & \mbox{\makebox[2.8em][r]{+0.72}\hspace{0.3em}{\scriptsize\textcolor{black!65}{[+0.08, +1.33]}}} & \mbox{\makebox[2.8em][r]{+0.42}\hspace{0.3em}{\scriptsize\textcolor{black!65}{[$-$0.20, +1.04]}}} \\
 & Llama-3.1-8B-It & {\footnotesize 49.70$\,\to\,$47.38$\,\to\,$48.24} & \mbox{\makebox[2.8em][r]{+0.47}\hspace{0.3em}{\scriptsize\textcolor{black!65}{[$-$0.17, +1.10]}}} & \mbox{\makebox[2.8em][r]{\textcolor{darkblue}{\textbf{+0.86}}}\hspace{0.3em}{\scriptsize\textcolor{black!65}{[+0.23, +1.48]}}} \\
 & Qwen3-32B & {\footnotesize 44.97$\,\to\,$42.80$\,\to\,$44.73} & \makebox[2.8em][r]{--} & \mbox{\makebox[2.8em][r]{\textcolor{darkblue}{\textbf{+1.94}}}\hspace{0.3em}{\scriptsize\textcolor{black!65}{[+1.22, +2.66]}}} \\
 & Qwen3.8-27B$^\ast$ & {\footnotesize 53.49$\,\to\,$53.82$\,\to\,$53.25} & \mbox{\makebox[2.8em][r]{$-$0.74}\hspace{0.3em}{\scriptsize\textcolor{black!65}{[$-$1.19, $-$0.30]}}} & \mbox{\makebox[2.8em][r]{$-$0.57}\hspace{0.3em}{\scriptsize\textcolor{black!65}{[$-$0.99, $-$0.17]}}} \\
\bottomrule
\end{tabular}
\caption{vLLM, 2.5 bits/value (FP8 scale and minimum, 80\,B/token/KV head), S4W128 on H200. Path: BF16 $\to$ KIVI-only $\to$ FlagRot-KIVI (per-key), scores 0--100. Gains: points over KIVI-only [nominal paired 95\% CI], same kernel and layout. Blue: CI above zero; --: not run. LongBench: 21-task macro-score, 4,750 items/model; RULER: 1,300 items; BF16-relative intervals: Table~\ref{tab:tasks25full}. Scores and contrasts rounded independently. $^\ast$Hybrid model: KV cache in 16 of 64 layers.}
\label{tab:tasks25}
\end{table*}

On long inputs, the lower error carries over to accuracy where KIVI-only loses it (Table~\ref{tab:tasks25}). In the
packed layout, selection adds \iTfRulerSel{} points to RULER at 128k tokens, about \iTfRulerRec\% of what KIVI-only loses, and \iTfThirtyTwoSel{} to the LongBench~v1 mean of the dense
Qwen3-32B, where KIVI-only falls \iTfThirtyTwoKiviLoss{} points below BF16 while \mkivi{} is not separated from
BF16 (\iTfThirtyTwoFr{} points). On the two 8B models the per-key gains are smaller, \iTfLbQSel{} and \iTfLbLSel{} LongBench points. The hybrid
Qwen3.8-27B keeps a KV cache in only 16 of its 64 layers, and there KIVI-only already matches BF16: selection does not
add to it (\iTfTwentySevenSel{} points), and \mkivi{} stays within \iTfTwentySevenFrAbs{} points of BF16. At equal bytes, \mkivi{} scores above \mauto{} on the three endpoints
where both ran, by \iPsRulerFa{} points on RULER (Table~\ref{tab:packed}). On those three paired endpoints,
the 2.5-bit and 3.0-bit scores differ by at most \iPsFpWorst{} points. In the 3.0-bit layout, selection adds \iRulerSelD{} RULER points, significant in our pre-specified Holm family, and
\iEoneSel{} LongBench points on Qwen3-32B (Table~\ref{tab:tasks2}).

Most of the selection gain needs no work at write time, as the quantization axis predicts: a KIVI group pools 32
tokens, so its best transform changes little from one group to the next (\S\ref{sec:attr}). In the fixed mode, \mkivi{} gains \iTfLbQFx{} and
\iTfLbLFx{} LongBench points on the 8B models and \iTfRulerFx{} on RULER, about as much as the per-key mode
(Table~\ref{tab:tasks25}). In the packed fidelity study its KL is \iFxKlStaticPk{} against \iUFkpTwo{}
for the per-key mode, which is lower on \iFxDynBetterPk{} of the 12 models.

\begin{table}[t]
\centering
\small
\setlength{\tabcolsep}{3pt}
\renewcommand{\arraystretch}{1.08}
\begin{tabular}{@{}p{\dimexpr0.31\columnwidth-6pt\relax}>{\columncolor{black!5}\raggedleft\arraybackslash}p{\dimexpr0.23\columnwidth-4pt\relax}>{\raggedleft\arraybackslash}p{\dimexpr0.22\columnwidth-4pt\relax}>{\raggedleft\arraybackslash}p{\dimexpr0.24\columnwidth-4pt\relax}@{}}
\toprule
 & \multicolumn{2}{c}{Equal-budget gain} & \\
\cmidrule(lr){2-3}
Endpoint & \shortstack[r]{vs. FlagRot-\\auto} & \shortstack[r]{vs.\\KIVI-only} & vs. BF16 \\
\midrule
Llama-It & \textcolor{darkblue}{\textbf{+18.5}} & \textcolor{darkblue}{\textbf{+2.9}} & $-$8.4 \\[-1pt]
RULER 128k & {\scriptsize [+16.5, +20.4]} & {\scriptsize [+1.2, +4.5]} & {\scriptsize [$-$10.0, $-$6.8]} \\[3pt]
Qwen3-8B & \textcolor{darkblue}{\textbf{+1.7}} & +0.4 & $-$1.6 \\[-1pt]
LongBench & {\scriptsize [+1.0, +2.4]} & {\scriptsize [$-$0.2, +1.0]} & {\scriptsize [$-$2.2, $-$1.1]} \\[3pt]
Llama-It & \textcolor{darkblue}{\textbf{+1.8}} & \textcolor{darkblue}{\textbf{+0.9}} & $-$1.5 \\[-1pt]
LongBench & {\scriptsize [+1.1, +2.5]} & {\scriptsize [+0.2, +1.5]} & {\scriptsize [$-$2.0, $-$0.9]} \\[3pt]
\bottomrule
\end{tabular}
\caption{2.5-bit \mkivi{} in vLLM: score differences in points [nominal paired 95\% CI]; all 2-bit caches at 80\,B/token/KV head. Blue: CI above zero. Llama-It: Llama-3.1-8B-Instruct.}
\label{tab:packed}
\end{table}

\begin{table}[t]
\centering
\small
\setlength{\tabcolsep}{2pt}
\begin{tabular*}{\columnwidth}{@{\extracolsep{\fill}}lrrr@{}}
\toprule
Group & INT4 ($-$\%) & MXFP4 ($-$\%) & NVFP4 ($-$\%) \\
\midrule
Dev. (4) & 55 {\scriptsize [51, 61]} & 69 {\scriptsize [65, 75]} & 46 {\scriptsize [38, 53]} \\
Larger (4) & 58 {\scriptsize [13, 69]} & 62 {\scriptsize [30, 69]} & 33 {\scriptsize [17, 56]} \\
Held-out (7) & 56 {\scriptsize [41, 57]} & 70 {\scriptsize [63, 89]} & 45 {\scriptsize [42, 55]} \\
\midrule
All (15) & 56 {\scriptsize [13, 69]} & 67 {\scriptsize [30, 89]} & 44 {\scriptsize [17, 56]} \\
\bottomrule
\end{tabular*}
\caption{4-bit caches without BF16 tokens: reduction of KL (\%) with \method{} relative to RTN, median over each group of models with the range in brackets. Significant (95\% CI) on 15 of 15 models per format. Qwen2.5-7B uses static key centering in both caches; per-model values are in Table~\ref{tab:kvmainfull}.}
\label{tab:kvmain}
\end{table}

\subsection{Per-Token Selection and Centering}
\label{sec:attr}
\begin{table}[t]
\centering
\small
\setlength{\tabcolsep}{2pt}
\begin{tabular*}{\columnwidth}{@{\extracolsep{\fill}}lrr@{}}
\toprule
\multicolumn{3}{@{}l}{\textbf{(a) LongBench v1: points added by selection}} \\
\midrule
Compared with & HF & vLLM \\
\midrule
\multicolumn{3}{@{}l}{\emph{Qwen3-8B}} \\
Centered RTN & \textbf{+3.70}\,{\scriptsize [+1.77, +5.66]} & \textbf{+3.81}\,{\scriptsize [+1.91, +5.75]} \\
Fixed mode & \textbf{+2.74}\,{\scriptsize [+1.01, +4.52]} & +1.48\,{\scriptsize [$-$0.15, +3.08]}$^\ddagger$ \\
\addlinespace[3pt]
\multicolumn{3}{@{}l}{\emph{Llama-3.1-8B-Instruct}} \\
Centered RTN & \textbf{+5.20}\,{\scriptsize [+3.27, +7.24]} & \textbf{+3.69}\,{\scriptsize [+1.71, +5.69]} \\
Fixed mode & \textbf{+2.42}\,{\scriptsize [+0.61, +4.26]} & +0.67\,{\scriptsize [$-$0.94, +2.37]}$^\ddagger$ \\
\midrule
\multicolumn{3}{@{}l}{\emph{Mean over models (Holm $p$)}} \\
Centered RTN & \textbf{+4.45}\,{\scriptsize ($<0.001$)} & \textbf{+3.75}\,{\scriptsize ($<0.001$)} \\
Fixed mode & \textbf{+2.58}\,{\scriptsize ($<0.001$)} & +1.07\,{\scriptsize (0.065)} \\
\bottomrule
\end{tabular*}
\par\vspace{5pt}
\begin{tabular*}{\columnwidth}{@{\extracolsep{\fill}}lrr@{}}
\toprule
\multicolumn{3}{@{}l}{\textbf{(b) AIME24+25: paired comparisons}} \\
\midrule
Comparison & Centering off & Centering on \\
\midrule
RTN $-$ BF16 & $-$34.8\,{\scriptsize [$-$43.1, $-$26.5]} & $-$17.5\,{\scriptsize [$-$23.5, $-$11.9]} \\
Per-key $-$ BF16 & $-$13.8\,{\scriptsize [$-$18.5, $-$9.2]} & $-$2.9\,{\scriptsize [$-$5.8, 0.0]} \\
\addlinespace[2pt]
Per-key $-$ RTN & \textbf{+21.0}\,{\scriptsize [+14.8, +27.5]} & \textbf{+14.6}\,{\scriptsize [+10.0, +19.8]} \\
\bottomrule
\end{tabular*}
\caption{Selection and centering on per-token INT2 caches (2.5-bit served layout; $g{=}32$, S4W128). Paired score differences [95\% CI]. (a) Per-key selection minus the named baseline; HF: HuggingFace simulation. (b) Qwen3-8B on H20. Bold: positive CIs or Holm-significant means. Experiment settings, $^\ddagger$fixed-baseline differences and AIME scores: Appendix~\ref{app:attr-settings}.}
\label{tab:attr2}
\end{table}

\begin{table}[t]
\centering
\small
\setlength{\tabcolsep}{3pt}
\begin{tabular*}{\columnwidth}{@{\extracolsep{\fill}}lrrr@{}}
\toprule
Variant & KL & Held-out & Worse \\
\midrule
\multicolumn{4}{@{}l}{\emph{\mauto{} (per-token keys)}} \\
\quad Full, weighted by $C_q$ & 17.8 & 17.8 & -- \\
\quad Plain reconstruction error & 29.2 & 30.4 & 12/12 \\
\quad Per-piece candidates only & 23.3 & 23.4 & 11/12 \\
\quad Whole-head candidate only & 21.0 & 21.2 & 12/12 \\
\quad No selection & 43.4 & 45.2 & 12/12 \\
\addlinespace[2pt]
\multicolumn{4}{@{}l@{}}{\makebox[0pt][l]{\emph{\mkivi{} (per-channel keys, 3.0-bit layout)}}} \\
\quad Full, attention-weighted & 12.3 & 11.8 & -- \\
\quad Weighted by $C_q$ only & 13.8 & 13.4 & 11/12 \\
\quad Plain reconstruction error & 19.1 & 19.1 & 12/12 \\
\quad No selection (KIVI) & 18.4 & 17.1 & 11/12 \\
\bottomrule
\end{tabular*}
\caption{Ablation at INT2 with S4W128: KL$\times10^3$, geometric mean over twelve models and over the eight held-out ones. Worse: models significantly above the full method (paired 95\% CI).}
\label{tab:ablation}
\end{table}

On per-token caches the decision has to follow the key. Per-key decisions agree with the fixed one on
\iMeAgreeTok\% of keys, against \iMeAgreeKivi\% of KIVI groups, and the per-key mode beats the fixed mode on the KL of
\iKlStaticWorse{} of 12 models and, in the HuggingFace study, on LongBench by \iFamHfFixD{} points after Holm correction
(Table~\ref{tab:attr2}a).
Centering and selection repair different errors: on AIME with thinking, centering recovers half of RTN's loss and
selection another \iAimeSelAutoD{} points (Table~\ref{tab:attr2}b).

Both the criterion and the candidates matter (Table~\ref{tab:ablation}). Judged by plain reconstruction error,
selection gives KIVI a KL of \iMeKivMse, against \iMeKivNone{} without selection, and dropping either kind of candidate
raises the KL of \mauto{}.

\subsection{The 4-bit Operating Point}
\label{sec:four}
\begin{table}[t]
\centering
\small
\setlength{\tabcolsep}{4pt}
\begin{tabular}{lrrr}
\toprule
 & \multicolumn{2}{c}{$\tau^2$, 3 domains} & SWE-V \\
\cmidrule(lr){2-3}\cmidrule(lr){4-4}
vs BF16 & pass\^{}1 & pass\^{}4 & resolved \\
\midrule
FP8 & \textcolor{gray}{$-$2.6} & \textcolor{gray}{$-$4.1} & \textcolor{gray}{$-$4.8} \\
 & {\scriptsize [$-$4.7, $-$0.5]} & {\scriptsize [$-$6.8, $-$1.5]} & {\scriptsize [$-$9.0, $-$0.6]} \\[1pt]
\method{}$^\star$ & $-$0.7 & \textcolor{gray}{$-$2.5} & $-$2.8 \\
 & {\scriptsize [$-$2.9, +1.6]} & {\scriptsize [$-$4.9, $-$0.1]} & {\scriptsize [$-$7.0, +1.4]} \\[1pt]
\midrule
BF16 rerun & $-$0.8 to +0.6 & $-$0.7 to $-$0.3 & $-$3.0 \\
 & & & {\scriptsize [$-$7.4, +1.4]} \\
\bottomrule
\end{tabular}
\caption{Qwen3.8-27B on $\tau^2$ (16 trials per task, 278 tasks pooled) and all 500 SWE-bench Verified instances: differences in points (95\% CI); gray: CI below zero. \emph{BF16 rerun}: for $\tau^2$, range over three 8-vs-8 seed splits; for SWE, a second run. $^\star$S4W128.}
\label{tab:agentic27}
\end{table}

Selection carries over to 4-bit caches, floating-point formats included. Without protected tokens, it lowers the KL
of INT4, MXFP4 and NVFP4 caches by a median of \kvMedInt\%, \kvMedMx\% and \kvMedNv\% over \nModels{} models,
significantly on each of them (Table~\ref{tab:kvmain}). At this width the cache keeps task accuracy, agentic tasks included: on Qwen3.8-27B,
its $\tau^2$-bench pass\textasciicircum1 and SWE-bench Verified scores stay within the spread of a BF16 rerun, while those
of FP8 fall below it (Table~\ref{tab:agentic27}).

\subsection{Cost}
\label{sec:serve}
\begin{table}[t]
\centering
\small
\setlength{\tabcolsep}{3pt}
\begin{tabular*}{\columnwidth}{@{\extracolsep{\fill}}lrrrr@{}}
\toprule
 & & \multicolumn{3}{c@{}}{tok/s at TPOT $\le$} \\
\cmidrule(l){3-5}
Cache & Seqs & 40\,ms & 60\,ms & 80\,ms \\
\midrule
\multicolumn{5}{@{}l}{\emph{Qwen3-8B}} \\
\quad BF16 & 23 & 707 & 707 & 707 \\
\quad FP8 & 46 & 1{,}494 & 1{,}494 & 1{,}494 \\
\quad \mauto{} & 142 & \textbf{1{,}905} & \textbf{2{,}150} & \textbf{2{,}175} \\
\addlinespace[2pt]
\multicolumn{5}{@{}l}{\emph{Qwen3-32B}} \\
\quad BF16 & 7 & 201 & 201 & 201 \\
\quad FP8 & 14 & 355 & 355 & 355 \\
\quad \mauto{} & 39 & \textbf{405} & \textbf{552} & \textbf{580} \\
\bottomrule
\end{tabular*}
\caption{Decoding throughput at a 32k-token context on one H200 (no prefix caching): generated tokens/s of the best batch within each time-per-output-token budget. Seqs: most 32k sequences that fit in the cache. \mauto{}: per-token INT2 at 2.5 bits per value.}
\label{tab:curve}
\end{table}

\begin{table}[t]
\centering
\small
\setlength{\tabcolsep}{2pt}
\begin{tabular*}{\columnwidth}{@{\extracolsep{\fill}}lrrrrrrr@{}}
\toprule
 & & \multicolumn{4}{c}{TPOT (ms) at batch} & Prefill & Cap. \\
\cmidrule(lr){3-6}
Cache & Bits & 64 & 32 & 8 & 1 & (tok/s) & (M) \\
\midrule
BF16 & 16\phantom{.0} & 29.1 & 22.0 & 17.1 & 14.9 & 12{,}031 & 0.93 \\
\addlinespace[2pt]
KIVI-only & 3.0 & 35.3 & 26.1 & 19.1 & 15.9 & 11{,}518 & 2.29 \\
FlagRot-KIVI & 3.0 & 40.7 & 29.0 & 19.7 & 16.0 & 10{,}893 & 2.29 \\
\addlinespace[2pt]
KIVI-only & 2.5 & 33.2 & 24.8 & 18.5 & 15.5 & 11{,}792 & 2.75 \\
FlagRot-KIVI & 2.5 & 38.7 & 27.8 & 19.1 & 15.7 & 11{,}172 & 2.75 \\
\addlinespace[2pt]
\shortstack[l]{Per-token\\(\mauto{})} & 2.5 & 30.5 & 23.5 & 18.4 & 16.5 & 11{,}024 & 2.75 \\
\bottomrule
\end{tabular*}
\caption{Pure decoding of Qwen3.8-27B in vLLM on one H200 (8192-token context, no prefix caching): time per output token, prefill throughput and KV capacity in tokens at equal memory. Caches at S4W128; KIVI rows at 2.5 bits store an FP8 scale and minimum; capacity as reported by vLLM, whose pages on this hybrid model also hold the linear-attention state; BF16 was re-measured in each session within 1\%.}
\label{tab:cost2}
\end{table}

Our vLLM plugin turns the smaller cache into capacity (Table~\ref{tab:cost2}). The packed layout holds \iCapPkEight{} times as many tokens as BF16 on Qwen3-8B; on the
hybrid Qwen3.8-27B, whose pages also carry linear-attention state, its attention pages hold \iCostPkPageX{} times as
many tokens and the whole cache \iCostPkCapX{} times. Where memory limits the batch, capacity becomes throughput: at
32k tokens, the per-token instance generates \iCvEightBf{} times the BF16 and \iCvEightFp{} times the FP8 token rate of
Qwen3-8B within a 60\,ms budget per token (Table~\ref{tab:curve}). The price is decode time at large batch: at batch 64,
the packed cache of the 27B model takes \iCostPkDec\% longer per output token than BF16. Per-key decisions also cost
\iCostPkPrefillSel\% of prefill throughput, which the fixed mode avoids.

\section{Related Work}
\label{sec:related}
KV-cache quantizers differ mainly in the axis they quantize and the tokens they protect. KIVI quantizes keys per channel
with a full-precision residual window \citep{liu2024kivi}, and KVQuant adds pre-RoPE keys, non-uniform codes and
sparse outliers \citep{hooper2024kvquant}; IntactKV and OTT keep pivot tokens intact
\citep{liu2024intactkv,su2025ott}. Other encoders use joint or vector codes
\citep{zandieh2025qjl,han2025polarquant,zhang2024cq}, store low-rank corrections of the residual
\citep{kang2024gear}, reorder channels \citep{duanmu2024skvq}, or vary the precision across tokens
\citep{he2024zipcache,yang2024mikv}.

Rotation methods fix their transforms during calibration or initialization. QuIP, QuIP\#, QuaRot and SpinQuant rotate
weights, activations or caches with fixed or learned rotations
\citep{chee2023quip,tseng2024quipsharp,ashkboos2024quarot,liu2025spinquant}. DuQuant constructs block rotations from
outlier dimensions, FlatQuant learns Kronecker-factored affine transforms, and KurTail fits a rotation that minimizes
kurtosis \citep{lin2024duquant,sun2025flatquant,akhondzadeh2025kurtail}. For KV caches, RotateKV reorders channels and
rotates groups of heads before RoPE \citep{rotatekv2025}, TurboQuant applies a random rotation before a scalar codebook
\citep{zandieh2026turboquant}, and WUSH-KV, OSCAR, OptR and AATC derive transforms or bit allocations from attention
statistics \citep{wushkv2026,oscar2026,optr2026,aatc2026}. All of these apply one transform to every key of a layer
or head. \method{} asks a different question, whether a piece should be transformed at all, and its decisions compose
with such transforms: on top of the transform fitted by WUSH-KV, or of RotateKV, selection lowers the KL of the
models we ran (Section~\ref{sec:fid}). Its fixed mode, like these methods, keeps one decision per head; its per-key mode
renews it without enlarging the cache.

Selecting a transform per block and signalling the selection is standard in media coding: video codecs pick among
transforms per block by rate--distortion cost \citep{sullivan2012hevc,zhao2021vvc}, and Four Over Six picks between two
FP4 scales per block \citep{fouroversix2025}. A KV cache changes both sides of that trade. The distortion is the error
attention sees, weighted by calibration queries and their softmax normalizers, and the side information hides in sign
bits the cache already stores. ScaNN weights quantization error by inner products with queries to choose codewords for
search \citep{guo2020scann}, much as GPTQ weights it by input statistics \citep{frantar2023gptq}; \method{} uses such a
weighting to select transforms, and lets flat pieces, which incoherence processing \citep{chee2023quip} would
rotate, stay as they are.

\section{Conclusion}
A low-bit KV cache need not commit to one rotation for all of its keys. We made the transform a decision of the
encoder, judged by the error the queries see, and showed that this lowers the error of the encoders we test and that
how often to decide follows the quantization axis. At an equal 2.5 bits per value, \mkivi{} has the lowest error among
the 2-bit caches we evaluate on most models and recovers part of the accuracy 2-bit KIVI loses on long inputs.

\section*{Limitations}
Our task evaluation covers four models from 8B to 32B parameters on NVIDIA H20 and H200 GPUs, and the fidelity study
twelve to fifteen open models. The task evaluation includes no mixture-of-experts model (the 4-bit fidelity study includes
Qwen3-30B-A3B), no model above 32B parameters and no other GPU generation, and our serving measurements use one inference engine. Other published low-bit caches enter our comparison
through simulated KL on twelve models, and RotateKV also through LongBench in simulation; served task comparisons cover
KIVI and our per-token instance. The selection statistics and the key mean are
calibrated on WikiText-2 by default; we test transfer to other text and one other calibration source. The shared key shift
preserves the softmax weights only for models without logit soft-capping, where calibration decides whether to apply it.

\section*{Ethics Statement}
The method compresses cached keys and values during inference with existing model weights and evaluation datasets, and
it introduces no new data. All models and datasets are publicly released and used under their licences.

\bibliography{references}

\appendix
\renewcommand{\topfraction}{0.9}
\renewcommand{\dbltopfraction}{0.9}
\renewcommand{\bottomfraction}{0.6}
\renewcommand{\textfraction}{0.07}
\renewcommand{\floatpagefraction}{0.7}
\renewcommand{\dblfloatpagefraction}{0.7}
\setcounter{topnumber}{4}
\setcounter{dbltopnumber}{3}
\setcounter{bottomnumber}{2}
\setcounter{totalnumber}{6}

\section{2-bit Caches: Details and Full Results}
\label{app:int2}
This appendix documents the INT2 experiments of \S\ref{sec:setup}--\S\ref{sec:serve}: the model cohort, the page layout
behind the bit counts, the agreement of the served kernels with the simulator, the RoPE-shifted calibration of fitted
transforms, and the per-model, per-task and serving results. Besides the names of the main text, the appendix uses three
labels for exploratory variants: \method{}-C and \mkivi{}-C run selection on top of a transform fitted on calibration
data, as WUSH-KV does, and \mlarge{} (4 bits only) selects from a larger candidate set of whole-head transforms.

\subsection{Models}
\label{app:int2-models}
The primary INT2 cohort holds the twelve models of Table~\ref{tab:permodel2}. Qwen3-32B enters the INT2 task study
(Table~\ref{tab:tasks25}) and a separate KL run (\S\ref{sec:tasks}), but not the twelve-model cohort, and Qwen2.5-7B is treated separately because of its key-projection biases
(Appendix~\ref{app:qwen25}). Table~\ref{tab:models} lists every model of the paper \citep{grattafiori2024llama3,touvron2023llama2,yang2025qwen3,qwen2024qwen25,qwen2026qwen38,jiang2023mistral,mistral2024nemo,gemma2024gemma2,gemma2025gemma3,abdin2024phi4,glm2024chatglm}. Llama-3.1-8B is the base model in
all KL results and the Instruct model in all task results.

\begin{table}[!htbp]
\centering
\small
\setlength{\tabcolsep}{2.6pt}
\begin{tabular}{@{}lrrrccc@{}}
\toprule
 & & & & QK- & & \\
Model & Layers & $H_{kv}$ & $D$ & norm & INT2 & Suite \\
\midrule
\multicolumn{7}{@{}l}{\emph{Development models}} \\
Llama-3.1-8B & 32 & 8 & 128 & no & \checkmark & \checkmark \\
Qwen3-8B & 36 & 8 & 128 & yes & \checkmark & \checkmark \\
Mistral-7B-v0.3 & 32 & 8 & 128 & no & \checkmark & \checkmark \\
Qwen3-14B & 40 & 8 & 128 & yes & \checkmark & \checkmark \\
\midrule
\multicolumn{7}{@{}l}{\emph{Larger models}} \\
Qwen3-32B & 64 & 8 & 128 & yes & -- & -- \\
Qwen3-30B-A3B & 48 & 4 & 128 & yes & -- & -- \\
Qwen3.8-27B & 16 of 64 & 4 & 256 & yes & \checkmark & \checkmark \\
Gemma-3-27B & 62 & 16 & 128 & yes & \checkmark & \checkmark \\
\midrule
\multicolumn{7}{@{}l}{\emph{Held-out models}} \\
Llama-2-7B & 32 & 32 & 128 & no & \checkmark & \checkmark \\
Llama-3.2-3B & 28 & 8 & 128 & no & \checkmark & \checkmark \\
Qwen2.5-7B & 28 & 4 & 128 & no & -- & \checkmark \\
Gemma-2-9B & 42 & 8 & 256 & no & \checkmark & \checkmark \\
Phi-4 & 40 & 10 & 128 & no & \checkmark & \checkmark \\
Mistral-NeMo-12B & 40 & 8 & 128 & no & \checkmark & \checkmark \\
GLM-4-9B & 40 & 2 & 128 & no & \checkmark & \checkmark \\
\bottomrule
\end{tabular}
\caption{Models: attention layers with a KV cache, KV heads $H_{kv}$ and head dimension $D$. All fifteen enter the
4-bit study (Table~\ref{tab:kvmain}). \emph{INT2}: the INT2 cohort; \emph{Suite}: the 4-bit baseline and
protected-token comparisons (Appendix~\ref{sec:kv}).}
\label{tab:models}
\end{table}

\subsection{Page Layout and Bits}
\label{app:int2-bits}
KIVI-only reserves the same flag space as \mkivi{}, so both use one page layout. For $D{=}128$, a token and KV head take
80 bytes in the per-token pages: two 32-byte code blocks and four 16-bit scale words each for keys and values. The
per-channel pages of \mkivi{} take 80 bytes in the packed layout: 64 bytes of codes, 8 bytes of FP8 key scales and
minima amortized over a 32-token group and 8 bytes of FP8 value scales and minima, with all flags in the sign bits of
the FP8 scales. In both layouts the five key flags of a group sit in the sign bits of the group's key scales and the
four per-piece value flags in the sign bits of the token's value scales; the packed layout keeps the value's whole-head
flag in the sign bit of a key-scale byte stored with the token. With FP16 scales and minima the pages take 97 bytes, 16 bytes each for keys and values plus
one byte for the value's whole-head flag; KIVI-only keeps this layout, so without selection it would need 96 bytes
(3.0 bits), with the same accuracy;
with $D{=}256$ the two layouts take 160 and 193 bytes. The BF16 copies of the
protected tokens, the BF16 keys of the incomplete key group and the kernels' workspace are not included. The
simulator's run counters charge the flags as 0.039 extra bits per value on each side, while the pages store them in
scale sign bits and, for the FP16 layout of \mkivi{}, one byte per token; the physical rates are 2.5 bits for the per-token and packed
per-channel pages and 3.03 bits for the FP16 layout.

\subsection{Kernels and Simulator Parity}
\label{app:int2-kernels}
In matched writer tests, the kernels reproduce the simulator's selections and codes. The served-to-simulated KL
ratio of the per-channel caches, in both layouts and both modes, lies between 0.96 and 1.03 on 20 to 146 windows of the
8B models and of Qwen3.8-27B, and CUDA graphs and eager execution generate identical sequences. On 60 windows of the base Llama-3.1-8B, the served KL$\times10^3$ is \iVklKivi{} for
KIVI-only and \iVklDyn{} for \mkivi{}; the fixed mode reaches \iVklStatic{} (FP16 layout). Table~\ref{tab:kernel2} times the decode attention of the two readers: the per-channel
reader takes \iKivDecFast{} times as long as the per-token one.

\begin{table}[!htbp]
\centering
\small
\setlength{\tabcolsep}{4pt}
\begin{tabular}{rrrrr}
\toprule
$B$ & Context & FlagRot-KIVI & FlagRot-auto & Ratio \\
\midrule
8 & 16k & 258 & 97 & 2.7 \\
8 & 32k & 467 & 162 & 2.9 \\
32 & 16k & 880 & 318 & 2.8 \\
32 & 32k & 1721 & 600 & 2.9 \\
64 & 16k & 1723 & 592 & 2.9 \\
64 & 32k & 3402 & 1161 & 2.9 \\
\bottomrule
\end{tabular}
\caption{Decode attention of one layer in $\mu$s (Llama-3.1-8B shapes, H20, CUDA graphs, median), FlagRot-KIVI in the packed 2.5-bit layout; \emph{Ratio}: FlagRot-KIVI over FlagRot-auto.}
\label{tab:kernel2}
\end{table}

\subsection{Calibrated Base and RoPE-Shifted Calibration}
\label{app:int2-calibrot}
The base-transform variants (\method{}-C per token and \mkivi{}-C per channel) fit invertible key, query and value
transforms per head and 32-channel block from second moments \citep{wushkv2026}. RoPE-shifted calibration mixes each
post-RoPE calibration statistic $S$ with the average of its RoPE-shifted copies,
\begin{equation*}
S\;\mapsto\;\tfrac12 S+\tfrac12\,\mathrm{mean}_{j}\,\R(o_j)\,S\,\R(o_j)^\top,
\end{equation*}
over the offsets $o_j=jL/8$, $j=1,\dots,7$, $L{=}32{,}768$; the shifted statistics reuse the original calibration
tokens. At 2k tokens it raises the twelve-model KL of \method{}-C from \iCrFc{} to \iCrFcRot{} and of \mkivi{}-C from
\iCrKc{} to \iCrKcRot; at 32k it lowers the KL of WUSH-KV's transform, with and without selection, on Qwen3-8B and
Llama-3.1-8B (Figure~\ref{fig:rdlen}d).
Under its published QuEST-2 protocol, WUSH-KV's own KL grows by \iWsGrowQwen$\times$ on both models over the same range
of lengths. Without RoPE-shifted calibration, a base fitted on 2k windows also drifts on 32k-token PG-19 books
(Table~\ref{tab:dom2}).

\subsection{Further Simulation Results}
\label{app:int2-sim}
Table~\ref{tab:fid2} details the INT2 comparison of Table~\ref{tab:unified}. Its first part adds the fixed mode on
centered per-token caches, the page bytes of the served layouts, RotateKV with 128-value groups, and for each cache the
number of models on which its KL is significantly below or above that of a reference. Its second part runs selection
under the protocol of WUSH-KV, on top of the transform that WUSH-KV fits, and its third part adds selection to RotateKV.
Table~\ref{tab:permodel2} gives the per-model KL.

\begin{table*}[t]
\centering
\small
\setlength{\tabcolsep}{3.5pt}
\begin{tabular}{@{}lrrl@{\hspace{16pt}}lrr@{}}
\toprule
\multicolumn{4}{@{}l}{(a) 12 models, INT2, 4 sink + 128 recent BF16 tokens} & \multicolumn{3}{l@{}}{(b) WUSH-KV setting, 12 models} \\
\cmidrule(r{16pt}){1-4}\cmidrule{5-7}
Cache & Bits & KL & Lower/higher & Cache & KL & L/H \\
\midrule
RTN & 2.5 & 75.4 &  & WUSH-KV transform & 35.7 &  \\
RTN, centered & 2.5 & 43.4 &  & \quad + selection on it & 26.7 & 12/0 \\
Fixed mode, centered & 2.5 & 27.4 & 0/11 FA & Head selection, centered & 24.8 & 11/0 \\
FlagRot-auto & 2.5 (80\,B) & 17.8 & 12/0 cRTN & FlagRot-auto & 78.8 & 1/11 \\
KIVI & 3.0 (97\,B) & 18.4 &  & & & \\
\cmidrule{5-7}
FlagRot-KIVI & 3.0 (97\,B) & 12.3 & 11/1 KIVI & \multicolumn{3}{@{}l@{}}{(c) RotateKV, 4 models, g32, clip 0.8} \\
KIVI, packed$^\S$ & 2.5 & 18.7 & 2/10 FA & S4W128 & 24.0 &  \\
FlagRot-KIVI, packed$^\S$ & 2.5 & 12.4 & 12/0 FA, 9/2 RKV & \quad + selection & 22.5 & 4/0 \\
RotateKV g32, clip 0.8 & 2.5 & 25.6 & 2/10 FA, 0/12 FK & W0 & 139 &  \\
RotateKV g32, clip 1.0 & 2.5 & 18.1 & 7/4 FA, 2/9 FK & \quad + selection & 127 & 4/0 \\
RotateKV g128, clip 0.8 & 2.125 & 22.3 & 6/5 FA, 0/12 FK &  & &  \\
\bottomrule
\end{tabular}
\caption{Simulated KL$\times10^3$ ($\downarrow$), geometric mean over models on all 2048-token WikiText-2 test windows. Bits: nominal bits per value with metadata (KIVI: FP16 scale and minimum; $^\S$FP8; others: 16 bits per group); in parentheses, page bytes per token and KV head in our kernels (97\,B: 3.03 bits at $D{=}128$). Lower/higher (L/H): models whose paired 95\% CI puts the KL below / above the reference (cRTN: centered RTN; FA: FlagRot-auto; FK: FlagRot-KIVI; RKV: RotateKV clip 1.0); the rest are not separated. (b)~Setting of Table~\ref{tab:wushkv} with QuEST-2, centering statistics from 4 of its FineWeb sequences; vs WUSH-KV. (c)~vs RotateKV.}
\label{tab:fid2}
\end{table*}

\begin{table*}[!tp]
\centering
\fontsize{8}{9.5}\selectfont
\setlength{\tabcolsep}{2.4pt}
\begin{tabular}{@{}l*{12}{r}@{}}
\toprule
 & \multicolumn{2}{c}{Per-token} & \multicolumn{3}{c}{Per-token, centered} & \multicolumn{2}{c}{Per-channel, 2.5 bits} & \multicolumn{2}{c}{Per-channel, 3.0 bits} & \multicolumn{3}{c}{RotateKV (our simulation)} \\
\cmidrule(lr){2-3}\cmidrule(lr){4-6}\cmidrule(lr){7-8}\cmidrule(lr){9-10}\cmidrule(l){11-13}
Model & RTN & FlagRot & RTN & Fixed & FlagRot-auto & KIVI & FlagRot-KIVI & KIVI & FlagRot-KIVI & g32, 0.8 & g32, 1.0 & g128, 0.8 \\
\midrule
Llama-3.1-8B & 115\phantom{.0} & 25.5 & 39.8 & 27.7 & 17.2 & 18.5 & \textbf{12.7} & 18.8 & 12.8 & 16.1 & 13.8 & 14.7 \\
Qwen3-8B & 83.9 & 44.7 & 54.9 & 39.6 & 25.7 & 29.8 & \textbf{18.6} & 30.0 & 19.0 & 43.0 & 28.0 & 29.0 \\
Mistral-7B-v0.3 & 44.5 & 13.7 & 33.9 & 21.8 & 13.8 & 17.2 & 9.3 & 17.0 & 9.3 & 14.9 & \textbf{8.9} & 11.1 \\
Qwen3-14B & 56.4 & 29.6 & 35.1 & 22.8 & 17.1 & 21.3 & \textbf{13.1} & 21.1 & 13.4 & 31.9 & 17.3 & 19.1 \\
Qwen3.8-27B & 27.0 & 21.1 & 22.5 & 19.3 & 18.7 & 19.6 & \textbf{14.2} & 20.0 & 14.3 & 26.2 & 20.6 & 21.8 \\
Gemma-3-27B & 459\phantom{.0} & 122\phantom{.0} & 286\phantom{.0} & 314\phantom{.0} & 73.8 & 29.7 & 40.7 & \textbf{25.1} & 34.9 & 163\phantom{.0} & 288\phantom{.0} & 273\phantom{.0} \\
Llama-2-7B & 37.3 & 12.0 & 32.1 & 15.9 & 11.6 & 13.5 & 7.4 & 12.5 & 7.3 & 17.2 & \textbf{6.7} & 10.1 \\
Llama-3.2-3B & 126\phantom{.0} & 31.9 & 84.9 & 30.8 & 19.4 & 22.7 & \textbf{14.6} & 24.0 & 14.6 & 20.1 & 17.5 & 18.5 \\
Gemma-2-9B & 20.8 & 6.9 & 14.1 & 6.3 & 6.1 & 7.2 & \textbf{4.2} & 7.3 & 4.2 & 10.6 & 4.7 & 6.2 \\
Phi-4 & 93.7 & 25.1 & 69.6 & 36.6 & 20.9 & 20.3 & 12.1 & 19.6 & \textbf{12.0} & 19.0 & 13.2 & 16.0 \\
Mistral-NeMo-12B & 44.8 & 18.1 & 39.0 & 24.2 & 17.8 & 22.7 & \textbf{13.4} & 22.3 & 13.5 & 18.9 & 13.5 & 15.9 \\
GLM-4-9B & 275\phantom{.0} & 199\phantom{.0} & 25.8 & 19.8 & 14.1 & 15.2 & 10.7 & 15.2 & \textbf{10.7} & 42.8 & 35.3 & 96.6 \\
\midrule
Geometric mean & 75.4 & 28.5 & 43.4 & 27.4 & 17.8 & 18.7 & 12.4 & 18.4 & \textbf{12.3} & 25.6 & 18.1 & 22.3 \\
\bottomrule
\end{tabular}
\caption{Per-model KL$\times10^3$ of INT2 caches at S4W128 (all WikiText-2 test windows). \emph{Centered}: selective key centering; per-channel 2.5 bits: FP8 scale and minimum; RotateKV columns: group size, clip ratio (2.5 bits for g32). Bold: lowest per model.}
\label{tab:permodel2}
\end{table*}

\paragraph{Gemma-3-27B.} This is the one model on which selection raises the INT2 KL of KIVI (Table~\ref{tab:permodel2}).
The cause is the query model of the criterion. In the \iGgLocal{} sliding-window layers, the sampled calibration
queries predict that selection lowers the logit error by \iGgCrit\%, while the queries that actually read those keys
see it rise by \iGgMeas\%; in the global layers prediction and measurement agree, as they do on the other models.
Evaluating the criterion with the real queries of each window, which a deployed encoder cannot do, gives a KL of
\iGgOracle{} against \iGmKivi{} for KIVI. A check on the calibration windows removes the regression without such
queries: it measures, per layer, the logit error that the calibration queries see under the selected and the plain
encoding, and turns selection off in layers where it is larger. The check turns off \iGgOff{} local layers of
Gemma-3-27B and at most \iGgOtherMax{} layers elsewhere (none on \iGgZero{} models). With it, Gemma-3-27B reaches
\iGgKl, \iGgVsKivi\% below KIVI, the other eleven models stay within their paired intervals, and the twelve-model
geometric mean moves from \iGgGmSel{} to \iGgGm. We added this check after observing the regression; all other results
are reported without it.

\paragraph{Joint selection.} Eq.~\ref{eq:select} chooses each piece of a key on its own. Searching all 16
combinations of the four pieces under the full criterion changes the KL of Llama-3.1-8B and Qwen3-8B by \iGmBestRelLlama\%
and \iGmBestRelQwen\%, both within their paired intervals, so the piece-wise choice loses little; on Gemma-3-27B, where the
calibration queries misjudge the local layers, the joint search follows them further and raises the KL by \iGmBestRel\%.

\paragraph{Calibration.} Table~\ref{tab:calibsens} varies the calibration of six models. \mkivi{} changes by at most
\iCsKiviMax\% in geometric mean. \mauto{} needs at least two windows: with one, its centering statistics raise the KL by
\iCsFaOne\%, and with two or more the change stays within \iCsFaRest\%.

\begin{table}[t]
\centering
\small
\setlength{\tabcolsep}{3pt}
\begin{tabular*}{\columnwidth}{@{\extracolsep{\fill}}lrrc@{}}
\toprule
Calibration variant & $\Delta$ (\%) & max & L/H/T \\
\midrule
\multicolumn{4}{@{}l}{\emph{\mkivi{} (3.0-bit layout)}} \\
1 window & +0.4 & 1.2 & 0/0/6 \\
2 windows & $-$0.7 & 1.5 & 3/0/3 \\
8 windows & $-$0.1 & 0.7 & 0/0/6 \\
16 windows & +0.2 & 1.8 & 0/0/6 \\
FineWeb-Edu instead of WikiText-2 & +2.1 & 3.7 & 0/4/2 \\
\quad same, tested on PG-19 & +1.3 & 3.5 & 0/2/4 \\
128 calibration queries & +1.7 & 2.4 & 0/4/2 \\
2048 calibration queries & $-$1.6 & 2.1 & 4/0/2 \\
Attention weight without the cap & +0.0 & 0.6 & 0/0/6 \\
Weight renormalized per window & $-$0.6 & 1.2 & 0/0/6 \\
Fixed mode, 16 windows$^\ast$ & $-$0.1 & 1.6 & 0/1/3 \\
\addlinespace[2pt]
\multicolumn{4}{@{}l}{\emph{\mauto{}}} \\
1 window & +5.8 & 19.6 & 0/3/3 \\
2 windows & +1.5 & 9.4 & 0/1/5 \\
8 windows & $-$1.8 & 5.3 & 4/0/2 \\
16 windows & $-$1.1 & 5.8 & 1/1/4 \\
FineWeb-Edu instead of WikiText-2 & +1.9 & 7.8 & 2/3/1 \\
\quad same, tested on PG-19 & +1.3 & 7.3 & 1/2/3 \\
\bottomrule
\end{tabular*}
\caption{Calibration sensitivity at INT2 with S4W128: change of the geometric-mean KL over six models (four development models, Llama-3.2-3B and Phi-4) relative to the default of four WikiText-2 windows, largest change on one model (max, \%), and models significantly lower / higher / tied (paired 95\% CI). $^\ast$Development models only.}
\label{tab:calibsens}
\end{table}

\paragraph{Centering and soft-capping.} On per-channel caches, centering every key group changes the simulated KL by
\iKiviCenterRange\% across the cohort without Gemma-2-9B. Gemma-2-9B soft-caps its attention logits at 50, and there the
same centering raises the KL of KIVI by \iSoftKivi\% and of \mkivi{} by \iSoftKiviSel\%.

\paragraph{Other encoders.} In an all-layer simulation of TurboQuant \citep{zandieh2026turboquant}, adding per-key
selection before its Hadamard transform changes the twelve-model KL at S4W128 from \iTqThree{} for 3-bit keys and values
and from \iTqFour{} for 3-bit keys and 4-bit values. The flags add 0.02 bits per value, averaged over keys and values.

\paragraph{WUSH-KV's published setting.} Table~\ref{tab:wushkv} follows WUSH-KV's published 4-bit configuration: the
QuEST quantizer \citep{panferov2025quest} with one scale per token and head, 16 sink and 128 recent tokens in BF16, and
calibration of every method on 128 FineWeb-Edu sequences \citep{penedo2024fineweb} of $\min(32\text{k}, \text{context
length})$ tokens; our implementation reproduces its published perplexities on Qwen3-8B (Table~\ref{tab:wushrepro}).
There, selection on top of WUSH-KV's own transform, with the whole-head candidate on keys and values, lowers the KL of WUSH-KV by \wsC\% on all \wsN{} models (geometric mean \wsCGm{} against \wsWushGm), and by
\wsCK\% when only keys are selected.

\begin{table}[!htbp]
\centering
\fontsize{8}{9.5}\selectfont
\setlength{\tabcolsep}{2.1pt}
\begin{tabular}{lrrrrrr}
\toprule
 & & & \multicolumn{4}{c}{\method{}} \\
\cmidrule(lr){4-7}
Model & Head & WUSH-KV & -H & -C$_K$ & -C & -C $8{\times}8$ \\
\midrule
Llama-3.1-8B & 2.03 & 1.78 & 1.37 & 1.53 & 1.44 & \textbf{1.36} \\
Qwen3-8B & 14.59 & 3.57 & 7.84 & 2.34 & 2.14 & \textbf{1.93} \\
Mistral-7B-v0.3 & 1.11 & 1.03 & \textbf{0.70} & 0.85 & 0.77 & 0.72 \\
Qwen3-14B & 2.99 & 2.61 & 1.74 & 1.72 & 1.56 & \textbf{1.42} \\
\midrule
Qwen3.8-27B & 5.08 & 4.70 & 3.38 & 3.62 & \textbf{3.16} & 3.26 \\
Gemma-3-27B & 13.02 & 3.41 & 3.59 & 2.52 & 2.40 & \textbf{2.25} \\
\midrule
Llama-2-7B & 1.20 & 1.09 & \textbf{0.82} & 0.94 & 0.87 & 0.82 \\
Llama-3.2-3B & 2.48 & 1.99 & 1.54 & 1.65 & 1.54 & \textbf{1.46} \\
Gemma-2-9B & 1.13 & 1.08 & 0.81 & 0.91 & 0.85 & \textbf{0.81} \\
Phi-4 & 2.04 & 1.71 & 1.30 & 1.44 & 1.35 & \textbf{1.27} \\
Mistral-NeMo-12B & 1.80 & 1.95 & \textbf{1.28} & 1.60 & 1.51 & 1.41 \\
GLM-4-9B & 11.02 & 11.84 & 3.60 & 2.27 & 2.20 & \textbf{1.88} \\
\midrule
Geo.\ mean & 3.13 & 2.33 & 1.78 & 1.63 & 1.51 & 1.41 \\
\bottomrule
\end{tabular}
\caption{WUSH-KV's published setting: QuEST 4-bit, one scale per token and head, S16W128, 128 FineWeb-Edu calibration sequences. KL$\times10^3$; bold: best. \method{}-C builds on WUSH-KV's transform; -C$_K$: keys only; -C $8{\times}8$: with \mlarge{} candidates. Qwen2.5-7B is excluded (all methods exceed 340).}
\label{tab:wushkv}
\end{table}

\paragraph{Rate and distortion on all models.} Table~\ref{tab:rd2} extends the points of Figure~\ref{fig:rdlen}a from
the development models to all twelve. Across code widths and group sizes, with and without BF16 tokens, selection
lowers the KL of RTN by \iRdTwFrVsRtn\% and that of KIVI by \iRdTwSelVsKivi\%, and \mkivi{} has the lowest KL of the
five caches at \iRdTwSelLowest{} point; at equal group size, the KIVI caches store 16 more metadata bits per group.

\begin{table}[!htbp]
\centering
\footnotesize
\setlength{\tabcolsep}{3.2pt}
\begin{tabular}{@{}l*{6}{r}@{}}
\toprule
 & \multicolumn{3}{c}{INT2} & \multicolumn{3}{c}{INT3} \\
\cmidrule(lr){2-4}\cmidrule(l){5-7}
Group size $g$ & 128 & 64 & 32 & 128 & 64 & 32 \\
\midrule
\multicolumn{7}{@{}l}{\emph{S4W128}} \\
RTN & 223\phantom{.0} & 106\phantom{.0} & 75.4 & 21.1 & 13.8 & 10.5 \\
FlagRot & 54.1 & 37.0 & 28.5 & 8.1 & 6.1 & 4.6 \\
FlagRot-C & 30.7 & 25.3 & 21.9 & 5.0 & 4.6 & 4.2 \\
KIVI & 44.7 & 28.4 & 18.4 & 5.8 & 4.4 & 3.2 \\
FlagRot-KIVI & 25.6 & 18.0 & 12.3 & 3.9 & 3.1 & 2.4 \\
\midrule
\multicolumn{7}{@{}l}{\emph{No BF16 tokens}} \\
RTN & 2603\phantom{.0} & 1529\phantom{.0} & 599\phantom{.0} & 329\phantom{.0} & 128\phantom{.0} & 69.5 \\
FlagRot & 731\phantom{.0} & 396\phantom{.0} & 218\phantom{.0} & 62.3 & 45.9 & 31.4 \\
FlagRot-C & 336\phantom{.0} & 235\phantom{.0} & 165\phantom{.0} & 37.6 & 31.7 & 25.2 \\
KIVI & 226\phantom{.0} & 136\phantom{.0} & 85.9 & 28.4 & 20.1 & 14.1 \\
FlagRot-KIVI & 105\phantom{.0} & 80.0 & 60.9 & 16.0 & 13.1 & 10.1 \\
\bottomrule
\end{tabular}
\caption{Rate and distortion: geometric-mean KL$\times10^3$ over twelve models. Bits per value: $b+16/g$ for per-token caches, $b+32/g$ for KIVI; without BF16 tokens, KIVI keeps the incomplete key group in BF16.}
\label{tab:rd2}
\end{table}

\paragraph{Other text.} Table~\ref{tab:dom2} covers all INT2 caches and lengths on other text: long
books, code, mathematical text and Chinese and German Wikipedia, with the calibration frozen on WikiText-2. At S4W128, \method{} and \mkivi{} lower the KL of RTN by
\iDomFr\% and \iDomFrk\% on each text and length, and \mkivi{} is below KIVI on \iDomFrkBelowKivi{} text.

\begin{table*}[!tp]
\centering
\fontsize{8}{9.5}\selectfont
\setlength{\tabcolsep}{3.1pt}
\begin{tabular}{@{}llrrrrrrrrrrr@{}}
\toprule
 & & & \multicolumn{5}{c}{Llama-3.1-8B} & \multicolumn{5}{c}{Qwen3-8B} \\
\cmidrule(lr){4-8}\cmidrule(lr){9-13}
 & & & KL & \multicolumn{4}{c}{$\Delta$ vs RTN (\%)} & KL & \multicolumn{4}{c}{$\Delta$ vs RTN (\%)} \\
\cmidrule(lr){5-8}\cmidrule(lr){10-13}
Text & Len. & Win. & RTN & FlagRot & FlagRot-C & KIVI & FlagRot-KIVI & RTN & FlagRot & FlagRot-C & KIVI & FlagRot-KIVI \\
\midrule
WikiText-2 & 2k & 141/146 & 115\phantom{.0} & \textbf{$-$78} & \textbf{$-$83} & \textbf{$-$84} & \textbf{$-$89} & 83.9 & \textbf{$-$47} & \textbf{$-$55} & \textbf{$-$64} & \textbf{$-$77} \\
PG-19 & 2k & 77 & 88.6 & \textbf{$-$77} & \textbf{$-$81} & \textbf{$-$89} & \textbf{$-$92} & 55.6 & \textbf{$-$46} & \textbf{$-$58} & \textbf{$-$67} & \textbf{$-$78} \\
PG-19 & 8k & 77 & 266\phantom{.0} & \textbf{$-$87} & \textbf{$-$89} & \textbf{$-$94} & \textbf{$-$96} & 132\phantom{.0} & \textbf{$-$44} & \textbf{$-$45} & \textbf{$-$80} & \textbf{$-$87} \\
PG-19 & 32k & 24 & 554\phantom{.0} & \textbf{$-$90} & \textbf{$-$70} & \textbf{$-$96} & \textbf{$-$97} & 292\phantom{.0} & \textbf{$-$28} & \textcolor{gray}{+366} & \textbf{$-$89} & \textbf{$-$92} \\
Code & 2k & 150 & 82.7 & \textbf{$-$73} & \textbf{$-$75} & \textbf{$-$80} & \textbf{$-$88} & 82.4 & \textbf{$-$45} & \textbf{$-$57} & \textbf{$-$70} & \textbf{$-$80} \\
Code & 8k & 80 & 163\phantom{.0} & \textbf{$-$78} & \textbf{$-$78} & \textbf{$-$86} & \textbf{$-$91} & 153\phantom{.0} & \textbf{$-$43} & \textbf{$-$43} & \textbf{$-$79} & \textbf{$-$86} \\
OpenWebMath & 2k & 150 & 99.2 & \textbf{$-$74} & \textbf{$-$78} & \textbf{$-$80} & \textbf{$-$87} & 86.2 & \textbf{$-$43} & \textbf{$-$63} & \textbf{$-$75} & \textbf{$-$83} \\
Wikipedia (zh) & 2k & 150 & 101\phantom{.0} & \textbf{$-$80} & \textbf{$-$85} & \textbf{$-$78} & \textbf{$-$88} & 62.8 & \textbf{$-$41} & \textbf{$-$60} & \textbf{$-$70} & \textbf{$-$80} \\
Wikipedia (de) & 2k & 150 & 88.8 & \textbf{$-$77} & \textbf{$-$79} & \textbf{$-$73} & \textbf{$-$84} & 64.1 & \textbf{$-$42} & \textbf{$-$61} & \textbf{$-$66} & \textbf{$-$77} \\
\bottomrule
\end{tabular}
\caption{INT2 caches at S4W128 on other text, calibration frozen on four WikiText-2 training windows: RTN KL$\times10^3$ and the change (\%) from it. \emph{Len./Win.}: window length and windows per model. Bold / gray: significantly below / above RTN (paired bootstrap, 95\% CI).}
\label{tab:dom2}
\end{table*}

\subsection{Task Results in vLLM}
\label{app:int2-tasks}
This subsection completes Table~\ref{tab:tasks25} with the remaining INT2 task results. The task settings are those of
\S\ref{sec:setup}, and caches are compared within one GPU type.

\begin{table*}[t]
\centering
\small
\setlength{\tabcolsep}{0pt}
\begin{tabular*}{\textwidth}{@{\extracolsep{\fill}}lllrrrrr@{}}
\toprule
 & & & & \multicolumn{2}{c}{$\Delta$ vs BF16} & \multicolumn{2}{c}{Selection over KIVI-only} \\
\cmidrule(lr){5-6}\cmidrule(l){7-8}
Model & Task & Items & BF16 & KIVI-only & \mkivi{} & Fixed & Per-key \\
\midrule
Qwen3-8B & LongBench v1 & 4{,}750 & 49.35 & $-$2.07 & $-$1.65 & +0.72 & +0.42 \\
 & & & & {\scriptsize [$-$2.69, $-$1.47]} & {\scriptsize [$-$2.19, $-$1.12]} & {\scriptsize [+0.08, +1.33]} & {\scriptsize [$-$0.20, +1.04]} \\[1pt]
Llama-3.1-8B-It & LongBench v1 & 4{,}750 & 49.70 & $-$2.32 & $-$1.46 & +0.47 & +0.86 \\
 & & & & {\scriptsize [$-$2.96, $-$1.70]} & {\scriptsize [$-$2.03, $-$0.91]} & {\scriptsize [$-$0.17, +1.10]} & {\scriptsize [+0.23, +1.48]} \\[1pt]
Qwen3-32B & LongBench v1 & 4{,}750 & 44.97 & $-$2.17 & $-$0.23 & -- & +1.94 \\
 & & & & {\scriptsize [$-$2.86, $-$1.49]} & {\scriptsize [$-$0.81, +0.36]} &  & {\scriptsize [+1.22, +2.66]} \\[1pt]
Qwen3.8-27B & LongBench v1 & 4{,}750 & 53.49 & +0.33 & $-$0.24 & $-$0.74 & $-$0.57 \\
 & & & & {\scriptsize [$-$0.07, +0.74]} & {\scriptsize [$-$0.59, +0.10]} & {\scriptsize [$-$1.19, $-$0.30]} & {\scriptsize [$-$0.99, $-$0.17]} \\[1pt]
Llama-3.1-8B-It & RULER 128k & 1{,}300 & 73.7 & $-$11.3 & $-$8.4 & +2.5 & +2.9 \\
 & & & & {\scriptsize [$-$12.9, $-$9.7]} & {\scriptsize [$-$10.0, $-$6.8]} & {\scriptsize [+0.9, +4.1]} & {\scriptsize [+1.2, +4.5]} \\[1pt]
\bottomrule
\end{tabular*}
\caption{\mkivi{} in vLLM at 2.5 bits per value (FP8 scale and minimum, 80\,B per token and KV head) with S4W128 on H200: BF16 score (\%) and paired differences in points with nominal 95\% CIs. Selection: \mkivi{} in its fixed and per-key mode minus KIVI-only, same kernel and layout; --: not run.}
\label{tab:tasks25full}
\end{table*}

Table~\ref{tab:tasks25full} retains the paired intervals relative to BF16 for the packed 2.5-bit results
summarized as score paths in Table~\ref{tab:tasks25}.

\paragraph{Fixed mode and per-key selection (3.0 bits).} Table~\ref{tab:fixed} gives both modes of \mkivi{} against KIVI-only on
the six endpoints where all three caches were run, with the same kernel, page layout and protection. The fixed mode
freezes one decision per layer, KV head and piece on the calibration windows; its simulated twelve-model KL is
\iFxKlStatic{}, between those of KIVI (\iKlKivi) and of the per-key mode (\iKlFrk).

\paragraph{3.0-bit layout.} Table~\ref{tab:tasks2} gives the task results of \mkivi{} with FP16 scales and minima.
They show the same broad task pattern as the packed layout of Table~\ref{tab:tasks25};
the pre-specified Holm family is defined on these 3.0-bit results.

\paragraph{Equal-budget per-channel and per-token caches.}
Table~\ref{tab:packed} compares the packed 2.5-bit per-channel cache with the 2.5-bit per-token cache on
three long-context endpoints, at 80 bytes per token and KV head. In paired comparisons against the 3.0-bit
per-channel cache, the packed cache differs by at most \iPsFpWorst{} points across these three endpoints.

\begin{table*}[t]
\centering
\small
\setlength{\tabcolsep}{3pt}
\renewcommand{\arraystretch}{1.15}
\begin{tabular}{@{}p{\dimexpr0.145\textwidth-3pt\relax}p{\dimexpr0.19\textwidth-6pt\relax}>{\raggedleft\arraybackslash}p{\dimexpr0.225\textwidth-6pt\relax}>{\raggedright\arraybackslash}p{\dimexpr0.195\textwidth-6pt\relax}>{\raggedright\arraybackslash}p{\dimexpr0.245\textwidth-3pt\relax}@{}}
\toprule
Benchmark & Model & {\footnotesize BF16 $\to$ KIVI $\to$ Ours} & \multicolumn{1}{c}{Fixed-mode gain} & \multicolumn{1}{c}{\textbf{Per-key gain}} \\
\midrule
RULER 128k & Llama-3.1-8B-It & {\footnotesize 73.7$\,\to\,$61.8$\,\to\,$65.9} & \mbox{\makebox[2.8em][r]{+3.9}\hspace{0.3em}{\scriptsize\textcolor{black!65}{[+2.3, +5.5]}}} & \mbox{\makebox[2.8em][r]{\textcolor{darkblue}{\textbf{+4.2}}$^\dagger$}\hspace{0.3em}{\scriptsize\textcolor{black!65}{[+2.6, +5.8]}}} \\
\addlinespace[4pt]
\multirow[t]{4}{*}{LongBench v1} & Qwen3-8B & {\footnotesize 49.35$\,\to\,$47.39$\,\to\,$48.45} & \mbox{\makebox[2.8em][r]{+0.5}\hspace{0.3em}{\scriptsize\textcolor{black!65}{[$-$0.1, +1.1]}}} & \mbox{\makebox[2.8em][r]{\textcolor{darkblue}{\textbf{+1.05}}}\hspace{0.3em}{\scriptsize\textcolor{black!65}{[+0.47, +1.64]}}} \\
 & Llama-3.1-8B-It & {\footnotesize 49.70$\,\to\,$47.55$\,\to\,$48.21} & \mbox{\makebox[2.8em][r]{+0.3}\hspace{0.3em}{\scriptsize\textcolor{black!65}{[$-$0.3, +1.0]}}} & \mbox{\makebox[2.8em][r]{\textcolor{darkblue}{\textbf{+0.66}}}\hspace{0.3em}{\scriptsize\textcolor{black!65}{[+0.03, +1.30]}}} \\
 & Qwen3.8-27B & {\footnotesize 53.49$\,\to\,$53.63$\,\to\,$53.18} & \mbox{\makebox[2.8em][r]{$-$1.0}\hspace{0.3em}{\scriptsize\textcolor{black!65}{[$-$1.4, $-$0.6]}}} & \mbox{\makebox[2.8em][r]{$-$0.45}\hspace{0.3em}{\scriptsize\textcolor{black!65}{[$-$0.89, 0.00]}}} \\
 & Qwen3-32B & {\footnotesize 44.97$\,\to\,$43.25$\,\to\,$44.65} & \makebox[2.8em][r]{--} & \mbox{\makebox[2.8em][r]{\textcolor{darkblue}{\textbf{+1.40}}}\hspace{0.3em}{\scriptsize\textcolor{black!65}{[+0.72, +2.07]}}} \\
\addlinespace[4pt]
AIME24+25 & Qwen3-8B & {\footnotesize 71.5$\,\to\,$68.5$\,\to\,$68.5} & \mbox{\makebox[2.8em][r]{+2.1}\hspace{0.3em}{\scriptsize\textcolor{black!65}{[$-$0.2, +4.4]}}} & \mbox{\makebox[2.8em][r]{0.0}\hspace{0.3em}{\scriptsize\textcolor{black!65}{[$-$2.7, +2.7]}}} \\
\addlinespace[4pt]
$\tau^2$-bench retail & Qwen3.8-27B & {\footnotesize 61.6$\,\to\,$61.6$\,\to\,$59.6} & \makebox[2.8em][r]{--} & \mbox{\makebox[2.8em][r]{$-$2.0}\hspace{0.3em}{\scriptsize\textcolor{black!65}{[$-$8.1, +4.4]}}} \\
\bottomrule
\end{tabular}
\caption{FlagRot-KIVI in vLLM, 3.0 bits/value (FP16 scale and minimum), S4W128 (H200; retail H20). Path: BF16 $\to$ KIVI-only $\to$ FlagRot-KIVI (per-key), scores 0--100. Gains: points over KIVI-only, same kernel [nominal paired 95\% CI]. Fixed: decided once from calibration data; per-key: decided as each key is written. Blue: CI above zero. $^\dagger$Holm (per-key) $p<0.001$ on RULER, $p=1$ on AIME/retail, one family; LongBench CIs descriptive. Metrics: RULER 13-subtask string-match mean (1,300 items); LongBench 21-task macro-score (4,750/model); AIME pass@1 (60 problems $\times$ 8 samples); retail DB-only pass$^1$ (114 tasks $\times$ 4 trials). Scores/contrasts rounded independently; --: not run.}
\label{tab:tasks2}
\end{table*}

\begin{table}[!htbp]
\centering
\fontsize{8}{9.5}\selectfont
\setlength{\tabcolsep}{3pt}
\begin{tabular}{@{}lrrr@{}}
\toprule
Endpoint & KIVI-only & Fixed & Per-key \\
\midrule
Llama-It, RULER 128k & 61.8 & +3.9 & +4.2 \\
 & & {\scriptsize [+2.3, +5.5]} & {\scriptsize [+2.5, +5.7]} \\[1pt]
Llama-It, LongBench & 47.5 & +0.3 & +0.7 \\
 & & {\scriptsize [$-$0.3, +1.0]} & {\scriptsize [+0.0, +1.3]} \\[1pt]
Qwen3-8B, LongBench & 47.4 & +0.5 & +1.1 \\
 & & {\scriptsize [$-$0.1, +1.1]} & {\scriptsize [+0.5, +1.6]} \\[1pt]
Qwen3.8-27B, LongBench & 53.6 & $-$1.0 & $-$0.5 \\
 & & {\scriptsize [$-$1.4, $-$0.6]} & {\scriptsize [$-$0.9, $-$0.0]} \\[1pt]
Qwen3-8B, AIME & 68.5 & +2.1 & +0.0 \\
 & & {\scriptsize [$-$0.2, +4.4]} & {\scriptsize [$-$2.7, +2.7]} \\[1pt]
Qwen3-8B, $\tau^2$ retail & 20.4 & $-$2.2 & +1.6 \\
 & & {\scriptsize [$-$6.2, +1.8]} & {\scriptsize [$-$2.0, +5.4]} \\[1pt]
\bottomrule
\end{tabular}
\caption{Fixed and per-key selection on the 3.0-bit per-channel cache in vLLM (FP16 scales and minima; H200, same kernel, layout and protection): score minus KIVI-only in points (nominal 95\% CI). Llama-It: Llama-3.1-8B-Instruct.}
\label{tab:fixed}
\end{table}

\paragraph{Selection and centering: task settings.}
\label{app:attr-settings}
Table~\ref{tab:attr2}(a) uses 16 LongBench v1 tasks and 800 items.
The fixed mode decides once from calibration data; the per-key mode decides as each key is written.
The HF study uses selective centering. The vLLM fixed baseline marked $^\ddagger$ instead centers
all key groups and fixes value transforms. Mean comparisons pool both models and use Holm correction;
the vLLM comparison against the fixed baseline does not reach $p<0.05$ ($p=0.065$).
Table~\ref{tab:attr2}(b) uses 60 AIME24+25 problems with eight samples each on Qwen3-8B and H20.
Its nominal paired 95\% intervals use resampling clustered by problem and are unadjusted.
Pass@1 (\%) is 72.3 for BF16. Enabling per-key selection changes the score from 37.5 to 58.5
without centering and from 54.8 to 69.4 with centering.

\paragraph{Per-token caches and agentic runs.} Table~\ref{tab:tasks2full} lists every INT2 task result of the
per-token caches, with RTN, uncentered selection and \mauto{} side by side. Table~\ref{tab:agentic2} gives the full
$\tau^2$-bench retail results of Qwen3-8B with 16 trials per task; its pooled rows
combine seed 300 with seeds 304--307.

\begin{table}[!htbp]
\centering
\footnotesize
\setlength{\tabcolsep}{2.2pt}
\begin{tabular}{@{}lrrrrr@{}}
\toprule
Cache & pass\textasciicircum1 & $\Delta$ & pass\textasciicircum4 & msg/sim & limit \\
\midrule
\multicolumn{6}{@{}l}{\emph{H200, 114 tasks $\times$ 16 trials}} \\
BF16 & 27.8 &  & 10.8 & 51.3 & 7.0\% \\
KIVI-only & 20.4 & $-$7.3 & 6.3 & 58.5 & 11.5\% \\
 & & {\scriptsize [$-$12.0, $-$2.7]} & & & \\
\mkivi{} & 22.1 & $-$5.7 & 5.4 & 62.0 & 12.3\% \\
 & & {\scriptsize [$-$10.4, $-$0.9]} & & & \\
Fixed mode & 18.2 & -- & -- & -- & -- \\
\mauto{} & 16.2 & $-$11.6 & 3.7 & 74.5 & 22.0\% \\
 & & {\scriptsize [$-$16.0, $-$7.4]} & & & \\
\midrule
\multicolumn{6}{@{}l}{\emph{H20, 114 tasks $\times$ 16 trials}} \\
BF16 & 26.8 &  & 10.9 & 50.9 & 6.9\% \\
\mauto{} & 17.4 & $-$9.4 & 4.1 & 72.6 & 22.1\% \\
 & & {\scriptsize [$-$14.1, $-$4.7]} & & & \\
\midrule
\multicolumn{6}{@{}l}{\emph{H20, 114 tasks $\times$ 20 trials, pooled}} \\
BF16 & 26.4 &  & 10.6 & 51.4 & 7.2\% \\
\mauto{} & 17.1 & $-$9.3 & 3.9 & 73.0 & 22.3\% \\
 & & {\scriptsize [$-$13.9, $-$4.9]} & & & \\
\bottomrule
\end{tabular}
\caption{Qwen3-8B on $\tau^2$-bench retail, INT2 caches at S4W128: database-state pass\textasciicircum1 and pass\textasciicircum4 (\%), paired $\Delta$ of pass\textasciicircum1 from BF16 (points, nominal 95\% CI), messages per simulation and share hitting the step limit. KIVI rows: 3.0-bit layout; fixed mode as in Table~\ref{tab:fixed}. Pooled: seeds 300, 304--307. --: not recorded.}
\label{tab:agentic2}
\end{table}

\begin{table}[!htbp]
\centering
\fontsize{8}{9.5}\selectfont
\setlength{\tabcolsep}{2.4pt}
\begin{tabular}{@{}lrrrrrr@{}}
\toprule
 & & \multicolumn{4}{c}{$\Delta$ vs BF16} & \\
\cmidrule(lr){3-6}
Task & BF16 & FP8 & RTN & FlagRot & -auto & $\Delta\Delta$ \\
\midrule
\multicolumn{7}{@{}l}{\emph{Qwen3-8B}} \\
LongBench v1$^\ddagger$ & 49.4 & +0.1 & $-$11.0 & $-$4.4 & $-$3.3 & +7.7 \\
LongBench v2 & 32.4 & 0.0 & $-$7.8 & $-$2.8 & $-$2.8 & +5.0 \\
RULER 32k & 73.2 & +0.7 & $-$32.7 & $-$10.7 & $-$8.8 & +23.9 \\
AIME24 & 75.0 & +0.4 & $-$35.4 & $-$9.6 & $-$1.7 & +33.7 \\
AIME25 & 69.6 & $-$2.5 & $-$34.2 & $-$17.9 & $-$4.2 & +30.0 \\
GPQA-Diamond & 58.4 & +0.2 & $-$13.5 & $-$0.5 & $-$2.7 & +10.8 \\
MATH-500 & 96.2 & +0.2 & $-$3.9 & $-$0.9 & $-$0.6 & +3.3 \\
GSM8K & 90.6 & $-$0.6 & $-$3.0 & $-$0.6 & +0.3 & +3.3 \\
MMLU-Pro & 61.8 & $-$0.2 & $-$10.4 & $-$7.0 & $-$3.6 & +6.8 \\
BFCL & 75.0 & +0.2 & $-$10.3 & $-$3.6 & $-$1.4 & +8.9 \\
$\tau^2$ airline, pass\textasciicircum1 & 16.5 & +3.0 & $-$1.5 & $-$4.0 & $-$0.5 & +1.0 \\
$\tau^2$ airline, pass\textasciicircum4 & 6.0 & +6.0 & +6.0 & $-$2.0 & $-$2.0 & $-$8.0 \\
$\tau^2$ retail, pass\textasciicircum1 & 24.8 & 0.0 & $-$19.7 & $-$11.8 & $-$9.2 & +10.5 \\
$\tau^2$ retail, pass\textasciicircum4 & 10.5 & $-$4.4 & $-$9.6 & $-$9.6 & $-$7.9 & +1.7 \\
$\tau^2$ telecom, pass\textasciicircum1 & 2.4 & $-$0.4 & 0.0 & $-$1.1 & +0.2 & +0.2 \\
$\tau^2$ telecom, pass\textasciicircum4 & 0.0 & 0.0 & +0.9 & 0.0 & 0.0 & $-$0.9 \\
SWE-bench Verified & 2.0 & $-$1.0 & $-$2.0 & $-$2.0 & $-$2.0 & 0.0 \\
\midrule
\multicolumn{7}{@{}l}{\emph{Llama-3.1-8B-It}} \\
LongBench v1$^\ddagger$ & 49.7 & $-$0.2 & $-$16.2 & $-$6.0 & $-$3.3 & +12.9 \\
LongBench v2 & 27.8 & +0.8 & +1.4 & +2.2 & +3.2 & +1.8 \\
RULER 32k & 86.1 & $-$1.2 & $-$56.9 & $-$15.7 & $-$7.2 & +49.7 \\
GSM8K & 85.8 & $-$2.2 & $-$8.9 & $-$0.8 & $-$0.2 & +8.7 \\
MMLU-Pro & 46.6 & +0.5 & $-$14.6 & $-$5.4 & $-$1.2 & +13.4 \\
BFCL & 31.3 & $-$0.5 & $-$6.7 & +0.7 & $-$1.9 & +4.8 \\
$\tau^2$ airline, pass\textasciicircum1 & 28.5 & $-$1.0 & $-$26.5 & +8.5 & $-$1.0 & +25.5 \\
$\tau^2$ airline, pass\textasciicircum4 & 20.0 & 0.0 & $-$20.0 & +10.0 & $-$2.0 & +18.0 \\
$\tau^2$ retail, pass\textasciicircum1 & 7.0 & +0.9 & $-$2.9 & $-$3.9 & $-$1.3 & +1.6 \\
$\tau^2$ retail, pass\textasciicircum4 & 3.5 & +0.9 & $-$2.6 & $-$2.6 & 0.0 & +2.6 \\
$\tau^2$ telecom, pass\textasciicircum1 & 7.0 & +2.4 & $-$6.1 & +5.5 & $-$7.0 & $-$0.9 \\
$\tau^2$ telecom, pass\textasciicircum4 & 7.0 & $-$3.5 & $-$7.0 & +3.5 & $-$7.0 & 0.0 \\
\midrule
\multicolumn{7}{@{}l}{\emph{Qwen3.8-27B}} \\
LongBench v1$^\ddagger$ & 53.5 & $-$0.1 & $-$1.0 & 0.0 & $-$0.2 & +0.7 \\
LongBench v2$^\ddagger$ & 50.7 & +0.4 & $-$3.6 & $-$2.2 & $-$3.4 & +0.2 \\
RULER 32k & 88.2 & $-$0.6 & $-$1.3 & +0.3 & $-$0.2 & +1.1 \\
AIME24 & 89.6 & $-$1.3 & $-$0.4 & $-$0.8 & $-$1.3 & $-$0.9 \\
AIME25 & 82.1 & +0.4 & +0.4 & $-$2.9 & $-$5.0 & $-$5.4 \\
GPQA-Diamond & 82.5 & $-$2.0 & $-$0.7 & $-$0.7 & $-$2.2 & $-$1.5 \\
MATH-500 & 98.5 & $-$0.3 & $-$0.1 & $-$0.5 & $-$0.2 & $-$0.1 \\
GSM8K & 96.4 & $-$0.2 & +0.1 & +0.2 & $-$0.1 & $-$0.2 \\
MMLU-Pro & 77.6 & $-$0.2 & $-$1.0 & $-$0.8 & $-$0.9 & +0.1 \\
$\tau^2$ airline, pass\textasciicircum1 & 62.0 & +3.0 & $-$14.5 & $-$0.5 & $-$6.5 & +8.0 \\
$\tau^2$ airline, pass\textasciicircum4 & 42.0 & +2.0 & $-$10.0 & $-$2.0 & $-$8.0 & +2.0 \\
$\tau^2$ retail, pass\textasciicircum1 & 61.6 & $-$1.3 & $-$4.6 & $-$2.4 & $-$2.0 & +2.6 \\
$\tau^2$ retail, pass\textasciicircum4 & 44.7 & $-$6.1 & $-$9.6 & $-$7.0 & $-$6.1 & +3.5 \\
$\tau^2$ telecom, pass\textasciicircum1 & 26.8 & $-$0.4 & $-$5.0 & $-$1.5 & $-$4.8 & +0.2 \\
$\tau^2$ telecom, pass\textasciicircum4 & 7.0 & $-$3.5 & $-$2.6 & $-$0.9 & $-$3.5 & $-$0.9 \\
SWE-bench Verified & 61.0 & $-$9.0 & $-$10.0 & $-$3.0 & $-$5.0 & +5.0 \\
\bottomrule
\end{tabular}
\caption{Per-token INT2 caches in vLLM on all tasks: BF16 score (\%) and paired differences in points. RTN, FlagRot and \emph{-auto} (FlagRot-auto) use S4W128; FP8: vLLM's FP8 cache with BF16 queries. $\Delta\Delta$: -auto minus RTN on the same items. $^\ddagger$H200 (21-task mean); all others H20.}
\label{tab:tasks2full}
\end{table}

\subsection{Serving}
\label{app:int2-serve}
\begin{figure}[!htbp]
\centering
\figpdf{\columnwidth}{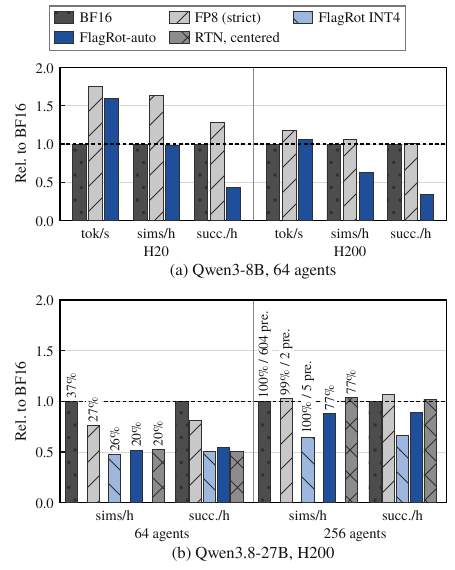}
\caption{Closed-loop $\tau^2$-bench retail with prefix caching and S4W128: generated tokens/s, completed and successful
simulations per hour, relative to BF16.}
\label{fig:serve}
\end{figure}

Figure~\ref{fig:serve} and Table~\ref{tab:serve2} report closed-loop $\tau^2$-bench retail serving with prefix caching.
On Qwen3.8-27B at
256 agents, the BF16 server preempts requests (\iQBfPre{} preemptions in its two runs) while the INT2 caches have none,
and \mauto{} completes simulations at \iQFaTwo$\times$ the BF16 rate and FP8 at \iQFpTwo$\times$. At 256 agents the BF16
cache of Qwen3-8B no longer holds all active conversations, on H200 and more so on the H20 with its smaller memory, so
the ratios at that load mainly reflect cache capacity.

\paragraph{Protocol.} Each client process drives 8 $\tau^2$-bench retail agents against one vLLM server with prefix
caching and a 512-sequence limit, and BF16 user simulators run on separate GPUs. Throughput counts generated tokens,
completed simulations and simulations with reward 1 over a timed window after warm-up. Qwen3-8B points at 64 and 128
agents have three runs for the main caches and two elsewhere, with a third run where the half-range of two exceeded
10\%, and points at 256 agents have two runs (one for KIVI-only on H200 and RTN on H20); we report the median of three
or the mean of two. Qwen3.8-27B has one run at 64 agents and two at 256.

\paragraph{Serving on a hybrid model.} For Qwen3.8-27B, vLLM~0.30 sizes each full-attention page to at least the state
page of the linear-attention layers, so the per-token INT2 format stores more tokens per page (\iQPage{} tokens for
BF16, FP8 and INT2).

\begin{table}[!htbp]
\centering
\fontsize{8}{9.5}\selectfont
\setlength{\tabcolsep}{1.6pt}
\begin{tabular}{@{}rlrrrrr@{}}
\toprule
Agents & Cache & tok/s & sims/h & succ./h & msg/sim & $\times$BF16 \\
\midrule
\multicolumn{7}{@{}l}{\emph{Qwen3-8B, H200}} \\
64 & BF16 & 1354 & 3257 & 861 & 49.3 & 1.00 \\
 & FP8 & 1598 & 3435 & 866 & 57.4 & 1.05 \\
 & FlagRot, INT4 & 1508 & 3211 & 843 & 53.3 & 0.99 \\
 & KIVI-only & 906 & 2153 & 387 & 53.8 & 0.66 \\
 & FlagRot-KIVI, fixed & 972 & 1892 & 369 & 61.2 & 0.58 \\
 & FlagRot-KIVI & 858 & 1415 & 260 & 70.5 & 0.43 \\
 & FlagRot-auto & 1436 & 2046 & 288 & 72.6 & 0.63 \\
 & RTN & 1773 & 1070 & 46 & 89.8 & 0.33 \\
\addlinespace[2pt]
128 & BF16 & 1637 & 3826 & 1010 & 51.4 & 1.00 \\
 & FP8 & 1816 & 3875 & 935 & 56.9 & 1.01 \\
 & FlagRot, INT4 & 1866 & 3777 & 1017 & 57.6 & 0.99 \\
 & FlagRot-auto & 1798 & 2140 & 323 & 77.0 & 0.56 \\
 & RTN & 2317 & 1077 & 50 & 89.5 & 0.28 \\
\addlinespace[2pt]
256 & BF16 & 558 & 1236 & 310 & 56.0 & 1.00 \\
 & FP8 & 2062 & 4645 & 1124 & 56.9 & 3.76 \\
 & FlagRot, INT4 & 2025 & 4290 & 1014 & 59.0 & 3.47 \\
 & KIVI-only & 1097 & 2449 & 430 & 58.2 & 1.98 \\
 & FlagRot-KIVI, fixed & 1140 & 2271 & 448 & 64.6 & 1.84 \\
 & FlagRot-KIVI & 1035 & 1672 & 324 & 70.9 & 1.35 \\
 & FlagRot-auto & 1737 & 2064 & 328 & 78.7 & 1.67 \\
 & RTN & 2150 & 1016 & 39 & 90.2 & 0.82 \\
\midrule
\multicolumn{7}{@{}l}{\emph{Qwen3-8B, H20}} \\
64 & BF16 & 414 & 957 & 274 & 51.1 & 1.00 \\
 & FP8 & 724 & 1567 & 352 & 57.2 & 1.64 \\
 & FlagRot, INT4 & 639 & 1362 & 309 & 55.5 & 1.42 \\
 & FlagRot-auto & 658 & 940 & 118 & 74.1 & 0.98 \\
 & RTN & 861 & 458 & 19 & 88.7 & 0.48 \\
\addlinespace[2pt]
128 & BF16 & 366 & 977 & 229 & 49.8 & 1.00 \\
 & FP8 & 771 & 1687 & 416 & 54.5 & 1.73 \\
 & FlagRot, INT4 & 742 & 1646 & 386 & 55.4 & 1.69 \\
\addlinespace[2pt]
256 & BF16 & 119 & 263 & 71 & 51.9 & 1.00 \\
 & FP8 & 805 & 1791 & 426 & 57.3 & 6.80 \\
 & FlagRot, INT4 & 730 & 1631 & 394 & 54.2 & 6.19 \\
 & FlagRot-auto & 656 & 891 & 139 & 74.7 & 3.38 \\
 & RTN & 670 & 304 & 15 & 94.0 & 1.16 \\
\midrule
\multicolumn{7}{@{}l}{\emph{Qwen3.8-27B, H200}} \\
64 & BF16 & 680 & 1917 & 1135 & 28.9 & 1.00 \\
 & FP8 & 549 & 1457 & 920 & 29.3 & 0.76 \\
 & FlagRot, INT4 & 335 & 901 & 577 & 28.1 & 0.47 \\
 & FlagRot-auto & 372 & 993 & 618 & 29.4 & 0.52 \\
 & RTN, centered & 377 & 1005 & 573 & 28.4 & 0.52 \\
\addlinespace[2pt]
256 & BF16 & 417 & 1114 & 654 & 31.1 & 1.00 \\
 & FP8 & 459 & 1146 & 698 & 30.5 & 1.03 \\
 & FlagRot, INT4 & 269 & 720 & 433 & 29.3 & 0.65 \\
 & FlagRot-auto & 408 & 982 & 579 & 30.6 & 0.88 \\
 & RTN, centered & 450 & 1153 & 668 & 30.1 & 1.04 \\
\bottomrule
\end{tabular}
\caption{Closed-loop $\tau^2$-bench retail serving with prefix caching; INT2 and INT4 caches at S4W128, KIVI rows in the packed 2.5-bit layout; FP8: vLLM's FP8 cache with BF16 queries. \emph{sims/h}, \emph{succ./h}: completed and successful simulations per hour; $\times$BF16: sims/h relative to BF16.}
\label{tab:serve2}
\end{table}

\paragraph{Pure decoding.} Without prefix caching and at a fixed context of 8192 tokens, the INT2 per-token cache of
Qwen3.8-27B takes \iCostDecPct\% more time per output token than BF16 at batch 64 (INT4: \iCostDecPctFour\%), and at the
same memory budget it holds \iCostCapX$\times$ as many tokens (\iCostCapTwo{} against \iCostCapBf{} million;
Table~\ref{tab:cost2}).

\section{HuggingFace Task Study}
\label{app:hf}
This appendix describes the HuggingFace study behind \S\ref{sec:attr} and Table~\ref{tab:attr2}(a), and the task-level
comparison with RotateKV cited in \S\ref{sec:fid}. The study runs Qwen3-8B and Llama-3.1-8B-Instruct with the simulated
caches in HuggingFace Transformers on 50 items of each of 16 English and code tasks of LongBench~v1, with the prompts of
the vLLM runs. On these items, the HF BF16 scores exceed the vLLM ones by \iHfVlQwen{} and \iHfVlLlama{} points.

In the HF study, the official and the protocol-matched RotateKV configurations produce very different task scores,
while the protected caches lie close together. Table~\ref{tab:hf2} lists the primary caches and controls. With S4W128,
RotateKV (g32, clip 0.8) scores below both \mauto{} and KIVI on Qwen3-8B and above both on Llama-3.1-8B-Instruct.
Without a window, in its official setting, it differs from BF16 by \iHfRkFaithQwen{} and \iHfRkFaithLlama{} points.
With clip ratio 1.0, RotateKV differs from BF16 by \iHfRkOneQwen{} and \iHfRkOneLlama{} points, and its direct
differences from \mauto{}, KIVI and \mkivi{} all have intervals that include zero.

\begin{table}[!htbp]
\centering
\footnotesize
\setlength{\tabcolsep}{3pt}
\begin{tabular}{@{}lrr@{}}
\toprule
Cache & Qwen3-8B & Llama-3.1-8B-It \\
\midrule
BF16 (score) & 49.90 & 50.13 \\
\midrule
\multicolumn{3}{@{}l}{\emph{Per-token INT2}} \\
RTN & $-$11.24 & $-$11.74 \\
RTN, centered & $-$5.92 & $-$6.97 \\
Fixed mode & $-$5.24 & $-$4.94 \\
Fixed mode, centered & $-$4.97 & $-$4.19 \\
FlagRot & $-$4.92 & $-$4.17 \\
FlagRot-auto & $-$2.23 & $-$1.77 \\
\midrule
\multicolumn{3}{@{}l}{\emph{Per-channel INT2}} \\
KIVI & $-$1.34 & $-$1.68 \\
FlagRot-KIVI & $-$1.49 & $-$1.28 \\
\midrule
\multicolumn{3}{@{}l}{\emph{RotateKV, INT2}} \\
g32, clip 0.8 & $-$3.01 & $-$0.63 \\
g32, clip 1.0 & $-$2.17 & $-$2.40 \\
official (W0) & $-$12.67 & $-$7.26 \\
\midrule
\multicolumn{3}{@{}l}{\emph{INT3, $g{=}128$}} \\
RTN, centered & $-$1.46 & $-$0.92 \\
FlagRot-auto & $-$1.11 & $-$0.53 \\
\bottomrule
\end{tabular}
\caption{HuggingFace study: BF16 score (\%) and differences from BF16 in points on 16 LongBench~v1 tasks (50 items each); INT2, $g{=}32$, S4W128 unless noted.}
\label{tab:hf2}
\end{table}

\FloatBarrier

\section{4-bit Caches}
\label{app:four}
\begin{table*}[!tp]
\centering
\fontsize{8}{9.5}\selectfont
\setlength{\tabcolsep}{1.8pt}
\begin{tabular}{lrrrrrrrrrrrrrrr}
\toprule
 & \multicolumn{5}{c}{INT4 ($g{=}32$)} & \multicolumn{5}{c}{MXFP4} & \multicolumn{5}{c}{NVFP4} \\
\cmidrule(lr){2-6}\cmidrule(lr){7-11}\cmidrule(lr){12-16}
Model & KL & Head & Fixed & FlagRot & -L & KL & Head & Fixed & FlagRot & -L & KL & Head & Fixed & FlagRot & -L \\
\midrule
\multicolumn{16}{l}{\emph{Development models}} \\
Llama-3.1-8B & 20.2 & $-$36.8 & $-$40.3 & $-$54.2 & \textbf{$-$72.7} & 95.1 & $-$70.7 & $-$68.3 & $-$72.7 & \textbf{$-$86.7} & 26.8 & $-$30.4 & $-$33.1 & $-$53.4 & \textbf{$-$72.1} \\
Qwen3-8B & 43.5 & $-$28.4 & $-$44.2 & $-$56.4 & \textbf{$-$73.2} & 140.3 & \textcolor{gray}{+41.5} & $-$49.9 & $-$65.0 & \textbf{$-$78.1} & 38.4 & \textcolor{gray}{+214.9} & $-$12.3 & $-$37.8 & \textbf{$-$59.2} \\
Mistral-7B-v0.3 & 7.1 & $-$33.6 & $-$37.6 & $-$51.0 & \textbf{$-$69.2} & 30.6 & $-$62.0 & $-$58.6 & $-$65.5 & \textbf{$-$83.3} & 10.6 & $-$35.2 & $-$32.6 & $-$51.1 & \textbf{$-$71.1} \\
Qwen3-14B & 25.1 & $-$46.3 & $-$49.0 & $-$61.4 & \textbf{$-$75.5} & 113.3 & $-$56.0 & $-$59.3 & $-$75.0 & \textbf{$-$86.0} & 24.2 & \textcolor{gray}{+31.6} & $-$12.8 & $-$40.2 & \textbf{$-$61.8} \\
\midrule
\multicolumn{16}{l}{\emph{Larger models}} \\
Qwen3-32B & 35.9 & $-$43.1 & $-$45.9 & $-$56.4 & \textbf{$-$70.2} & 139.7 & $-$52.0 & $-$56.1 & $-$68.8 & \textbf{$-$81.3} & 34.9 & \textcolor{gray}{+26.3} & $-$12.2 & $-$34.7 & \textbf{$-$54.3} \\
Qwen3-30B-A3B & 53.3 & $-$30.4 & $-$48.1 & $-$59.5 & \textbf{$-$74.1} & 140.4 & $-$18.9 & $-$46.3 & $-$59.7 & \textbf{$-$77.3} & 47.9 & \textcolor{gray}{+57.0} & $-$4.8 & $-$32.2 & \textbf{$-$51.5} \\
Qwen3.8-27B & 6.7 & $-$8.0 & $-$10.3 & $-$13.3 & \textbf{$-$34.2} & 15.7 & $-$20.9 & $-$29.9 & $-$29.9 & \textbf{$-$51.3} & 8.3 & $-$3.4 & $-$6.9 & $-$16.8 & \textbf{$-$37.4} \\
Gemma-3-27B & 63.1 & $-$26.8 & $-$34.8 & $-$68.8 & \textbf{$-$87.7} & 113.6 & $-$6.3 & $-$32.1 & $-$64.7 & \textbf{$-$85.6} & 43.2 & \textcolor{gray}{+82.2} & $-$10.4 & $-$56.1 & \textbf{$-$80.3} \\
\midrule
\multicolumn{16}{l}{\emph{Held-out models}} \\
Llama-2-7B & 9.2 & $-$47.1 & $-$47.0 & $-$56.1 & \textbf{$-$72.6} & 60.8 & $-$80.0 & $-$79.6 & $-$79.6 & \textbf{$-$90.8} & 12.8 & $-$43.9 & $-$43.5 & $-$55.3 & \textbf{$-$73.1} \\
Llama-3.2-3B & 23.8 & $-$36.6 & $-$42.0 & $-$56.5 & \textbf{$-$74.9} & 87.1 & $-$61.1 & $-$59.5 & $-$67.2 & \textbf{$-$83.6} & 30.6 & $-$26.1 & $-$29.2 & $-$52.2 & \textbf{$-$72.0} \\
Qwen2.5-7B & 4697.5 & \textcolor{gray}{+19.7} & \textcolor{gray}{+14.0} & \textcolor{gray}{+6.2} & \textbf{$-$49.4} & 6215.7 & \textcolor{gray}{+3.7} & \textcolor{gray}{+11.7} & \textcolor{gray}{+10.7} & \textbf{$-$28.6} & 5074.7 & \textcolor{gray}{+31.6} & \textcolor{gray}{+22.3} & \textcolor{gray}{+20.1} & \textbf{$-$5.2} \\
Gemma-2-9B & 5.6 & $-$41.0 & $-$42.3 & $-$40.9 & \textbf{$-$60.0} & 22.6 & $-$69.9 & $-$69.6 & $-$69.5 & \textbf{$-$81.5} & 7.4 & $-$37.8 & $-$38.9 & $-$45.1 & \textbf{$-$62.1} \\
Phi-4 & 14.3 & $-$37.9 & $-$40.9 & $-$56.8 & \textbf{$-$71.7} & 40.6 & $-$52.7 & $-$52.1 & $-$63.3 & \textbf{$-$78.8} & 18.5 & $-$30.0 & $-$31.3 & $-$50.9 & \textbf{$-$69.6} \\
Mistral-NeMo-12B & 9.5 & $-$34.4 & $-$38.2 & $-$45.2 & \textbf{$-$64.4} & 45.6 & $-$68.7 & $-$66.2 & $-$67.4 & \textbf{$-$84.3} & 12.4 & $-$26.3 & $-$25.8 & $-$42.3 & \textbf{$-$64.0} \\
GLM-4-9B & 42.9 & $-$20.4 & $-$37.6 & $-$56.1 & \textbf{$-$73.9} & 521.6 & $-$83.8 & $-$86.0 & $-$88.7 & \textbf{$-$94.3} & 64.0 & $-$7.8 & $-$11.1 & $-$44.2 & \textbf{$-$71.3} \\
\bottomrule
\end{tabular}
\caption{4-bit KV caches without BF16 tokens: KL$\times10^3$ without rotation and relative KL change (\%) of each rotation. \emph{Head}: whole-head Hadamard; \emph{Fixed}: fixed mode; \emph{-L}: \mlarge{}. Bold: best and significantly below zero; gray: significantly above zero (95\% CI). Qwen2.5-7B without key centering.}
\label{tab:kvmainfull}
\end{table*}

\begin{table*}[!tp]
\centering
\fontsize{8}{9.5}\selectfont
\setlength{\tabcolsep}{2.2pt}
\begin{tabular}{lrrrrrrrrrrrrrr}
\toprule
 & \multicolumn{9}{c}{INT4} & \multicolumn{5}{c}{NVFP4} \\
\cmidrule(lr){2-10}\cmidrule(lr){11-15}
Model & RTN & TurboQ & Smooth & Head & Base & Fixed & FlagRot & -L & -C & RTN & 4/6 & 4/6+FR & FlagRot & -L \\
\emph{bits/elem.} & \emph{4.50} & \emph{4.9--7.1} & \emph{4.50} & \emph{4.50} & \emph{4.50} & \emph{4.50} & \emph{4.50} & \emph{4.62} & \emph{4.50} & \emph{4.50} & \emph{4.50} & \emph{4.50} & \emph{4.50} & \emph{4.71} \\
\midrule
Llama-3.1-8B & 20.2 & 16.6 & 18.4 & 12.7 & 7.8 & 12.0 & 9.2 & \textbf{5.5} & 7.0 & 26.8 & 21.0 & 10.5 & 12.5 & \textbf{7.5} \\
Qwen3-8B & 43.5 & 27.3 & 33.2 & 31.2 & 14.8 & 24.3 & 19.0 & \textbf{11.7} & 13.3 & 38.4 & 33.4 & 19.3 & 23.9 & \textbf{15.7} \\
Mistral-7B-v0.3 & 7.1 & 6.0 & 6.7 & 4.7 & 3.2 & 4.4 & 3.5 & \textbf{2.2} & 2.8 & 10.6 & 8.4 & 4.3 & 5.2 & \textbf{3.1} \\
Qwen3-14B & 25.1 & 17.2 & 21.2 & 13.5 & 10.9 & 12.8 & 9.7 & \textbf{6.1} & 9.7 & 24.2 & 20.1 & 12.0 & 14.5 & \textbf{9.2} \\
\midrule
Qwen3.8-27B & 6.7 & 8.2 & 6.2 & 6.2 & 5.2 & 6.0 & 5.8 & \textbf{4.4} & 5.1 & 8.3 & 7.2 & 6.1 & 6.9 & \textbf{5.2} \\
Gemma-3-27B & 63.1 & 72.3 & 32.5 & 46.2 & 10.8 & 41.1 & 19.7 & \textbf{7.8} & 9.8 & 43.2 & 37.9 & 17.0 & 19.0 & \textbf{8.5} \\
\midrule
Llama-2-7B & 9.2 & 6.5 & 8.9 & 4.9 & 4.6 & 4.9 & 4.0 & \textbf{2.5} & 3.9 & 12.8 & 10.5 & 4.9 & 5.7 & \textbf{3.4} \\
Llama-3.2-3B & 23.8 & 20.9 & 22.5 & 15.1 & 9.2 & 13.8 & 10.4 & \textbf{6.0} & 8.0 & 30.6 & 24.9 & 12.2 & 14.6 & \textbf{8.6} \\
Qwen2.5-7B & 4697.5 & \textbf{17.9} & 3768.9 & 5621.3 & 4611.7 & 5354.1 & 4989.7 & 2378.9 & 4285.1 & 5074.7 & 6123.9 & 5950.3 & 6095.9 & \textbf{4812.4} \\
Gemma-2-9B & 5.6 & 7.1 & 5.6 & 3.3 & 3.5 & 3.2 & 3.3 & \textbf{2.2} & 3.2 & 7.4 & 6.1 & 3.6 & 4.1 & \textbf{2.8} \\
Phi-4 & 14.3 & 11.9 & 14.5 & 8.9 & 5.4 & 8.5 & 6.2 & \textbf{4.1} & 5.0 & 18.5 & 15.3 & 7.6 & 9.1 & \textbf{5.6} \\
Mistral-NeMo-12B & 9.5 & 8.0 & 9.5 & 6.2 & 5.2 & 5.9 & 5.2 & \textbf{3.4} & 4.7 & 12.4 & 10.2 & 6.0 & 7.1 & \textbf{4.5} \\
GLM-4-9B & 42.9 & 32.2 & 32.2 & 34.2 & 15.2 & 26.8 & 18.8 & \textbf{11.2} & 13.4 & 64.0 & 51.9 & 28.3 & 35.7 & \textbf{18.4} \\
\bottomrule
\end{tabular}
\caption{Recent KV-cache quantizers without BF16 tokens: KL$\times10^3$; bold: best per group. \emph{TurboQ}: TurboQuant in vLLM; \emph{Smooth}: SmoothAttention; \emph{Head}: whole-head Hadamard; \emph{Base}: calibrated base alone; \emph{-L}/\emph{-C}: \mlarge{}/\method{}-C; \emph{4/6}: Four Over Six; \emph{FR}: \method{}. Qwen2.5-7B without key centering.}
\label{tab:kvsota}
\end{table*}

\begin{table*}[!tp]
\centering
\small
\setlength{\tabcolsep}{3pt}
\begin{tabular}{lrrrrrrrrr}
\toprule
 & \multicolumn{5}{c}{$S{=}4$, $W{=}128$} & \multicolumn{4}{c}{$W{=}128$ only} \\
\cmidrule(lr){2-6}\cmidrule(lr){7-10}
Model & RTN & Fixed & \method{} & \method{}-C & \mlarge{} & RTN & Fixed & \method{} & \method{}-C \\
\midrule
Llama-3.1-8B & 2.30 & 1.52 & 1.31 & 1.28 & 1.05 & 3.22 & 1.92 & 1.83 & 1.71 \\
Qwen3-8B & 4.05 & 2.30 & 2.09 & 1.93 & 1.60 & 10.00 & 5.01 & 3.96 & 7.14 \\
Mistral-7B-v0.3 & 1.34 & 0.78 & 0.70 & 0.71 & 0.52 & 1.46 & 0.83 & 0.77 & 0.79 \\
Qwen3-14B & 2.71 & 1.69 & 1.40 & 1.46 & 1.25 & 7.36 & 3.74 & 2.78 & 5.56 \\
Qwen3.8-27B & 3.38 & 2.92 & 2.88 & 2.92 & 2.53 & 3.80 & 3.34 & 2.89 & 3.46 \\
Gemma-3-27B & 8.14 & 6.73 & 3.24 & 2.56 & 2.04 & 14.76 & 12.97 & 5.64 & 4.31 \\
Llama-2-7B & 1.33 & 0.87 & 0.82 & 0.85 & 0.65 & 1.43 & 0.90 & 0.88 & 0.90 \\
Llama-3.2-3B & 2.79 & 1.81 & 1.51 & 1.44 & 1.16 & 3.87 & 2.31 & 2.17 & 1.88 \\
Qwen2.5-7B & 659.94 & 4029.46 & 3468.10 & 3483.43 & 896.79 & 743.62 & 4066.84 & 3513.18 & 3530.62 \\
Gemma-2-9B & 1.18 & 0.82 & 0.78 & 0.81 & 0.68 & 1.70 & 1.05 & 1.08 & 1.30 \\
Phi-4 & 2.43 & 1.49 & 1.26 & 1.16 & 0.98 & 2.56 & 1.54 & 1.33 & 1.28 \\
Mistral-NeMo-12B & 1.97 & 1.32 & 1.20 & 1.27 & 0.97 & 2.35 & 1.48 & 1.43 & 1.47 \\
GLM-4-9B & 10.79 & 8.50 & 2.90 & 2.95 & 2.14 & 19.97 & 12.55 & 6.54 & 4.60 \\
\midrule
Geo.\ mean, $S{=}4$, $W{=}0$ & 15.17 & 9.30 & 7.17 & 5.30 & 4.51 & & & & \\
Geo.\ mean, $S{=}4$, $W{=}32$ & 4.26 & 2.66 & 2.12 & 1.92 & 1.59 & & & & \\
Geo.\ mean, $S{=}4$, $W{=}256$ & 2.07 & 1.54 & 1.19 & 1.22 & 0.94 & & & & \\
\bottomrule
\end{tabular}
\caption{Per-model KL$\times10^3$ of INT4 caches with BF16 sink and recent tokens; geometric means for other windows exclude Qwen2.5-7B (no key centering; every per-token 4-bit cache exceeds 600).}
\label{tab:swfull}
\end{table*}

\begin{table*}[!tp]
\centering
\small
\setlength{\tabcolsep}{4pt}
\begin{tabular}{lrlrlrl}
\toprule
 & \multicolumn{2}{c}{Qwen3-8B} & \multicolumn{2}{c}{Llama-3.1-8B-It} & \multicolumn{2}{c}{Qwen3.8-27B} \\
\cmidrule(lr){2-3}\cmidrule(lr){4-5}\cmidrule(lr){6-7}
Task & BF16 & $\Delta$ \method{} [95\% CI] & BF16 & $\Delta$ \method{} [95\% CI] & BF16 & $\Delta$ \method{} [95\% CI] \\
\midrule
RULER (32k) & 73.2 & $-$0.1 {\scriptsize [$-$1.1, +0.9]} & 86.1 & $-$0.2 {\scriptsize [$-$0.8, +0.3]} & 88.2 & +0.1 {\scriptsize [$-$0.1, +0.3]} \\
LongBench v2 & 32.4 & +0.4 {\scriptsize [$-$1.0, +1.8]} & 27.8 & \textbf{+1.4} {\scriptsize [+0.2, +2.6]} & 50.7 & $-$0.6 {\scriptsize [$-$3.2, +2.2]} \\
LongBench v1$^\P$ & 49.2 & 0.0 {\scriptsize [$-$0.3, +0.3]} & 49.6 & $-$0.1 {\scriptsize [$-$0.5, +0.2]} & 54.3 & $-$0.2 {\scriptsize [$-$0.4, +0.1]} \\
AIME24 & 75.0 & $-$2.5 {\scriptsize [$-$6.7, +1.2]} & -- & -- & 89.6 & +2.5 {\scriptsize [$-$2.1, +8.3]} \\
AIME25 & 69.6 & $-$4.6 {\scriptsize [$-$11.2, +1.2]} & -- & -- & 82.1 & +1.3 {\scriptsize [$-$1.7, +4.2]} \\
GPQA-Diamond & 58.4 & +2.2 {\scriptsize [$-$0.7, +5.2]} & -- & -- & 82.5 & $-$0.5 {\scriptsize [$-$2.5, +1.5]} \\
MATH-500 & 96.2 & +0.4 {\scriptsize [$-$0.4, +1.2]} & -- & -- & 98.5 & 0.0 {\scriptsize [$-$0.5, +0.5]} \\
GSM8K & 90.6 & $-$0.5 {\scriptsize [$-$1.3, +0.2]} & 85.8 & +0.2 {\scriptsize [$-$0.9, +1.4]} & 96.4 & $-$0.1 {\scriptsize [$-$0.6, +0.5]} \\
MMLU-Pro & 61.8 & +0.1 {\scriptsize [$-$1.3, +1.6]} & 46.6 & +0.1 {\scriptsize [$-$1.6, +1.9]} & 77.6 & $-$0.1 {\scriptsize [$-$1.4, +1.1]} \\
BFCL & 75.0 & $-$0.3 {\scriptsize [$-$1.1, +0.6]} & 31.3 & \textbf{+0.9} {\scriptsize [+0.5, +1.4]} & -- & -- \\
$\tau^2$ airline & 16.5$^\dagger$ & 0.0 {\scriptsize [$-$5.0, +4.5]} & 28.5 & +1.0 {\scriptsize [$-$5.0, +6.5]} & 62.0 & $-$4.5 {\scriptsize [$-$14.0, +4.0]} \\
$\tau^2$ retail$^\ddagger$ & 24.8 & 0.0 {\scriptsize [$-$5.5, +5.9]} & 7.0$^\dagger$ & +0.9 {\scriptsize [$-$1.5, +3.5]} & 61.6 & $-$3.3 {\scriptsize [$-$8.1, +1.8]} \\
$\tau^2$ telecom & 2.4$^\dagger$ & $-$0.2 {\scriptsize [$-$2.0, +1.3]} & 7.0$^\dagger$ & 0.0 {\scriptsize [+0.0, +0.0]} & 26.8 & $-$2.4 {\scriptsize [$-$7.9, +3.1]} \\
SWE-bench Verified$^\S$ & 2.0$^\dagger$ & 0.0 {\scriptsize [$-$4.0, +4.0]} & -- & -- & 61.0 & $-$6.0 {\scriptsize [$-$15.0, +3.0]} \\
\bottomrule
\end{tabular}
\caption{\method{} (INT4, S4W128) in vLLM: BF16 score (\%) and paired $\Delta$ from BF16 in points. Bold / gray: CI above / below zero. $^\dagger$BF16 below 20\%. $^\ddagger$Database-state reward. $^\S$100-instance subset. $^\P$Mean of 16 English tasks on H200, with its own BF16 run. --: not evaluated.}
\label{tab:tasks}
\end{table*}

\begin{table*}[!tp]
\centering
\small
\setlength{\tabcolsep}{5pt}
\begin{tabular}{lrrrrrr}
\toprule
 & \multicolumn{3}{c}{H200} & \multicolumn{3}{c}{H20} \\
\cmidrule(lr){2-4}\cmidrule(lr){5-7}
KV cache \hfill concurrent agents & 64 & 128 & 256 & 64 & 128 & 256 \\
\midrule
BF16$^\dagger$ & 1381 & 1402 & 637 & 482 & 166 & 134 \\
FP8 (FlashAttention-3)$^\dagger$ & 1468 & 1608 & 1744 & 589 & 654 & 706 \\
vLLM INT4$^\dagger$ & 391 & 484 & 577 & 256 & 288 & 320 \\
FP8 (FlashInfer) & 1660 & 2051 & 2197 & 768 & 819 & 994 \\
\method{}, no BF16 tokens & 1808 & 2051 & 2398 & 737 & 781 & 852 \\
\midrule
RTN$^\star$ (INT4, our kernels) & 1672 & 1972 & 2277 & 737 & 775 & 869 \\
Fixed mode$^\star$ & 1509 & 1819 & 2190 & 623 & 741 & 831 \\
\method{}$^\star$ & 1523 & 1889 & 2062 & 703 & 783 & 829 \\
\method{}-C$^\star$ & 1269 & 1554 & 1720 & 644 & 683 & 732 \\
\bottomrule
\end{tabular}
\caption{Serving $\tau^2$-bench retail agents with Qwen3-8B and INT4 caches: generated tokens/s. $^\star$S4W128. $^\dagger$Earlier run block of the same protocol.}
\label{tab:tput}
\end{table*}

At 4 bits, selection lowers the KL on the 15 models and three formats we test, while task scores differ little
between encoders. This appendix gives the 4-bit encoder and its evaluation, the fidelity results behind \S\ref{sec:four}, the
task study and the serving kernels; the mechanism is analyzed in Appendix~\ref{app:mech}.

\subsection{Setup and Encoder Settings}
\label{sec:method4}
At 4 bits, \method{} uses the per-token key candidates of \mauto{} and the query-weighted criterion of
\S\ref{sec:select} on INT4 (asymmetric, $g{=}32$), MXFP4 and NVFP4 caches. Each key head of dimension $D$ is split into
sub-vectors of $r_b{=}32$ channels (16 for NVFP4), so every block transform stays inside one scale group. For a new
token, every sub-vector selects between the identity and $\had_{r_b}$, and per head this selection competes with the
whole-head transform $\had_D$. The selection takes $D/r_b+1$ flag bits per key head (5 for $D{=}128$ and $r_b{=}32$),
stored in the sign bits of the token's INT4 or NVFP4 scales, which are always positive: one per key scale group, and
the head flag in a value scale. INT4 stores its zero point in the low mantissa bits of the FP16 scale, so the cache
keeps 4.5 bits per element; MXFP4, whose scales have no sign bit, stores the flags separately. Values are quantized
without transform (Appendix~\ref{sec:mech}).

\paragraph{Selection criterion.} A key error $\Delta$ changes the logit of a query $q$ by $q\cdot\Delta$. We weight it
with the second moment of the post-RoPE queries of the head's GQA group, $C_q=\mathbb{E}[qq^\top]$, estimated once on
calibration data: $\varepsilon_k(\Delta)=\Delta^\top C_q\Delta$, with $\Delta$ the reconstruction error in the original
coordinates. The encoder (i)~quantizes the key three times, unrotated, with $\had_{r_b}$ per sub-vector and with
$\had_D$; (ii)~gives every sub-vector the candidate with the lower error under its diagonal block of $C_q$;
(iii)~keeps the assembled reconstruction or the $\had_D$ one, whichever has the lower error under the full $C_q$. Where
values are selected, their error is weighted by the column energy of $W_o$. For keys the inverse transform is moved to
the query as in Eq.~\eqref{eq:attn}, so attention never materializes the dequantized keys.

\paragraph{Variants.} The \emph{fixed mode} fixes the selection of every key and value head on calibration data, per
sub-vector or for the whole head, and stores no flags. \method{}-C selects on top of a calibrated base transform: per
head and scale group, invertible matrices $B$ for keys and $B_v$ for values are derived in closed form from second
moments of keys, queries and output weights \citep{wushkv2026}. The cache stores $\hat c=Q(T_jBk)$ for the selected
candidate $T_j$; queries are multiplied by $B^{-\top}$, so that $q^\top B^{-1}T_j^\top\hat c$ approximates $q^\top k$,
and the attention output by $B_v^{-1}$. \mlarge{} selects on keys and values
among eight candidates per sub-vector (the identity, $\had_{r_b}$ and six $\had_{r_b}S_j$) and eight per head (the
per-sub-vector selection, $\had_D$ and six $\had_D S_j$), with 15 flag bits per 128 values (27 with $r_b{=}16$ for
NVFP4); the $S_j$ are diagonal $\pm1$ matrices drawn once from a seeded generator.

\paragraph{Models, calibration and windows.} The development models are Llama-3.1-8B, Qwen3-8B, Qwen3-14B and
Mistral-7B-v0.3; all design decisions were taken on their WikiText-2 validation windows. We also evaluate Qwen3-32B,
Qwen3-30B-A3B, Qwen3.8-27B, Gemma-3-27B and seven held-out models (Table~\ref{tab:models}). Keys and values are quantized
after RoPE with linear layers in BF16; in Qwen3.8-27B only the full-attention layers, which hold the KV cache, are
quantized. The selection statistics come from four 2048-token windows of WikiText-2 \emph{train}. The metric is
KL(BF16$\|$Q)$\times10^3$, averaged over all positions of all 2048-token windows of the WikiText-2 test set
\citep{merity2017wikitext}; for Gemma-2 and Gemma-3 every window starts with BOS followed by 2047 text tokens. Without
protected tokens, every query reads all earlier tokens, including those of its own window, from the quantized cache;
S4W128 keeps the first 4 and the 128 most recent tokens in BF16. The simulation keeps full-precision INT4 scales and zero
points; storing them in the deployed format changes KL by $+0.7\%$ on average. Intervals are paired percentile-bootstrap
95\% intervals over windows (5000 resamples; \citealp{efron1993bootstrap}), and a difference is called significant when
its interval excludes zero and the effect is at least 1\%.

\subsection{Fidelity Across Formats}
\label{sec:kv}
With static key centering on Qwen2.5-7B, \method{} lowers KL by \kvFourBitRangeAll\% relative to round-to-nearest (RTN)
on the 15 models and three formats without protected tokens (Table~\ref{tab:kvmain}). Table~\ref{tab:kvmainfull} adds
whole-head rotation (QuaRot-style), the fixed mode and \mlarge{}, without centering: the KL of \method{} is below that of
whole-head rotation in \dynWins{} model--format pairs, and \mlarge{} lowers KL by \kvLargeRangeAll\% on the models other than Qwen2.5-7B in each format.

\paragraph{Recent quantizers.} Table~\ref{tab:sotacond} compares recent per-token KV quantizers on 12 models
(per model in Table~\ref{tab:kvsota}). At 4.5 bits per element, \method{} lowers KL by \oursVsSmooth\% relative to
SmoothAttention \citep{lin2025qserve} and by \oursVsTurbo\% relative to vLLM's TurboQuant setting
\citep{zandieh2026turboquant}, which keeps the first and last two layers in BF16 and uses \turboBitsRange{} bits per
element. On NVFP4 it lowers the KL of RTN by \nvOursVsRtn\% and composes with the scale selection of Four Over Six
\citep{fouroversix2025}, lowering its KL by \nvPlusOurs\%. The calibrated base of WUSH-KV alone has a lower mean KL than
\method{} (\sotaGmCalib{} against \sotaGmOurs{}); selection on top of the base, \method{}-C, lowers the KL of the base by
a further \baseCOverW\%. Table~\ref{tab:wushkv} applies selection in WUSH-KV's own published setting
(Appendix~\ref{app:int2-sim}).

\begin{table}[!htbp]
\centering
\small
\setlength{\tabcolsep}{4pt}
\begin{tabular}{lrr}
\toprule
Method & bits & KL \\
\midrule
\multicolumn{3}{l}{\emph{INT4}} \\
RTN & 4.50 & 16.72 \\
TurboQuant (vLLM) & 4.9--7.1 & 14.43 \\
SmoothAttention & 4.50 & 14.53 \\
Whole-head Hadamard & 4.50 & 11.05 \\
Calibrated base & 4.50 & 7.02 \\
Fixed mode & 4.50 & 10.15 \\
\method{} & 4.50 & 7.88 \\
\mlarge{} & 4.62 & \textbf{4.81} \\
\method{}-C & 4.50 & 6.30 \\
\midrule
\multicolumn{3}{l}{\emph{NVFP4}} \\
RTN & 4.50 & 19.96 \\
Four Over Six & 4.50 & 16.53 \\
Four Over Six + \method{} & 4.50 & 9.04 \\
\method{} & 4.50 & 10.72 \\
\mlarge{} & 4.71 & \textbf{6.49} \\
\bottomrule
\end{tabular}
\caption{Recent KV quantizers without BF16 tokens: geometric-mean KL$\times10^3$ over 12 models; \emph{bits}: per cached element. Per model: Table~\ref{tab:kvsota}.}
\label{tab:sotacond}
\end{table}

\paragraph{Key centering on Qwen2.5-7B.}
\label{app:qwen25}
On Qwen2.5-7B, the key-projection biases of the first and last layers add a constant offset of several hundred to a few
low-frequency channels, which then sets the scale of every group that contains them; the uncentered per-token 4-bit
caches reach a KL of thousands ($\qcRtnPlain\times10^{-3}$ for RTN). Static key centering subtracts the calibrated
per-channel mean of the post-RoPE keys from every cached key, including the BF16 ones. This changes all logits of a
query by the same amount, so it leaves the unquantized softmax weights unchanged and needs no per-token metadata. With
centering, \method{} lowers the KL of RTN by \qcOursVsRtn\% without protected tokens and by \qcOursVsRtnSw\% at S4W128
across the three formats. We apply centering only to Qwen2.5-7B, whose biases cause the offset; on the other models its
effect varies in sign.

\begin{table}[!htbp]
\centering
\small
\setlength{\tabcolsep}{5pt}
\begin{tabular}{lrrrr}
\toprule
BF16 sinks $S$ & 0 & 0 & 4 & 16 \\
BF16 window $W$ & 0 & 128 & 128 & 128 \\
\midrule
RTN & 16.72 & 4.09 & 2.78 & 2.70 \\
Fixed mode & 10.15 & 2.56 & 1.88 & 1.87 \\
\method{} & 7.88 & 2.08 & 1.48 & 1.46 \\
\method{}-C & 6.30 & 2.23 & 1.45 & 1.42 \\
\midrule
\mlarge{} & 4.81 & 1.46 & 1.16 & 1.14 \\
\midrule
FP8 cache, per tensor & 3.71 & & & \\
\bottomrule
\end{tabular}
\caption{INT4 caches with $S$ sink and $W$ recent tokens in BF16: geometric-mean KL$\times10^3$ over 12 models (all but Qwen2.5-7B); per model in Table~\ref{tab:swfull}.}
\label{tab:window}
\end{table}

\paragraph{Protected tokens.} Table~\ref{tab:window} combines selection with BF16 sink and recent tokens, and
Table~\ref{tab:swfull} gives the per-model values. For \method{}, the geometric-mean KL falls from \swOursNone{} without
protection to \swOursWin{} with a 128-token window and to \swOursDef{} with four sinks added; 16 sinks lower it by a
further \swSixteenGain\%. At S4W128, the geometric-mean KL of \method{} over the 12 models is \swOursVsRtn\% lower than
that of RTN and \swOursVsStatic\% lower than that of the fixed mode under the same protection, and it is below that of
an unprotected FP8 cache with one scale per tensor (\swOursDef{} against \swFp). The task and serving comparisons below
use S4W128.

\subsection{Task Accuracy}
\label{app:tasks}
\begin{table}[!htbp]
\centering
\small
\setlength{\tabcolsep}{1.9pt}
\begin{tabular}{lrrr}
\toprule
Cache & Qwen3-8B & Llama-3.1-8B-It & Qwen3.8-27B \\
\midrule
BF16 & 49.15 & 49.62 & 54.27 \\
FP8 & $-$0.1 & \textcolor{gray}{$-$0.7} & +0.1 \\
 & {\scriptsize [$-$0.6, +0.3]} & {\scriptsize [$-$1.3, $-$0.2]} & {\scriptsize [$-$0.1, +0.3]} \\[1pt]
vLLM INT4 & \textcolor{gray}{$-$1.0} & \textcolor{gray}{$-$1.1} & +0.3 \\
 & {\scriptsize [$-$1.6, $-$0.5]} & {\scriptsize [$-$1.7, $-$0.4]} & {\scriptsize [$-$0.1, +0.6]} \\[1pt]
RTN$^\star$ & +0.2 & \textcolor{gray}{$-$0.6} & 0.0 \\
 & {\scriptsize [$-$0.1, +0.6]} & {\scriptsize [$-$1.0, $-$0.1]} & {\scriptsize [$-$0.2, +0.2]} \\[1pt]
Fixed mode$^\star$ & 0.0 & $-$0.1 & +0.1 \\
 & {\scriptsize [$-$0.3, +0.4]} & {\scriptsize [$-$0.5, +0.2]} & {\scriptsize [$-$0.2, +0.3]} \\[1pt]
\method{}$^\star$ & 0.0 & $-$0.1 & $-$0.2 \\
 & {\scriptsize [$-$0.3, +0.3]} & {\scriptsize [$-$0.5, +0.2]} & {\scriptsize [$-$0.4, +0.1]} \\[1pt]
\method{}-C$^\star$ & $-$0.1 & $-$0.3 & $-$0.1 \\
 & {\scriptsize [$-$0.4, +0.2]} & {\scriptsize [$-$0.7, +0.0]} & {\scriptsize [$-$0.3, +0.2]} \\[1pt]
\method{} & +0.1 & \textcolor{gray}{$-$0.6} & +0.1 \\
 & {\scriptsize [$-$0.4, +0.6]} & {\scriptsize [$-$1.1, $-$0.1]} & {\scriptsize [$-$0.2, +0.4]} \\[1pt]
\midrule
vs RTN$^\star$ & $-$0.3 & \textbf{+0.5} & $-$0.2 \\
 & {\scriptsize [$-$0.6, +0.1]} & {\scriptsize [+0.0, +1.0]} & {\scriptsize [$-$0.4, +0.0]} \\
\bottomrule
\end{tabular}
\caption{LongBench v1 (16 English tasks), INT4 caches on H200: BF16 score and paired differences in points (95\% CI, items resampled within tasks). Bold / gray: CI above / below zero. $^\star$S4W128; \emph{vs RTN$^\star$}: \method{}$^\star$ minus RTN$^\star$.}
\label{tab:lb1}
\end{table}

The task study holds the served model and the items fixed and changes only the KV cache. We serve Qwen3-8B,
Llama-3.1-8B-Instruct and Qwen3.8-27B in vLLM \citep{kwon2023vllm} with BF16 weights on one GPU per server, with BF16,
FP8 and vLLM's INT4 cache and with RTN, the fixed mode, \method{} and \method{}-C at S4W128 (marked $^\star$). The tasks
cover long-context retrieval and understanding (RULER at 32k tokens \citep{hsieh2024ruler}, 13 subtasks with 100 items
each; LongBench v1 \citep{bai2024longbench}, all 21 tasks with contexts truncated in the middle to 31.5k tokens;
LongBench v2 \citep{bai2025longbenchv2}, all 503 items truncated to 32k tokens), reasoning with long generations (AIME
2024/2025 with 8 samples per problem, GPQA-Diamond \citep{rein2024gpqa} and MATH-500 \citep{lightman2024verify} with 3,
using the models' recommended sampling), short tasks (GSM8K \citep{cobbe2021gsm8k}, MMLU-Pro \citep{wang2024mmlupro}),
tool use (BFCL \citep{patil2025bfcl}, all 4441 items; $\tau^2$-bench \citep{barres2025tau2}, airline, retail and telecom
with 4 trials per task and a BF16 Qwen3-8B user simulator) and coding (a fixed 100-instance subset of SWE-bench Verified
\citep{jimenez2024swebench} with mini-SWE-agent). Each task uses identical items, prompts and sampling for all caches;
tasks without a sampling setting above decode greedily, and unparseable answers count as wrong. Samples of an item are
averaged before pairing, and intervals are percentile bootstrap over items (2000 resamples). Within a task, every cache and its BF16 reference run
on the same GPU type; LongBench v1 runs on H200.

Table~\ref{tab:tasks} reports \tkBOursN{} model--task pairs: \tkBOursIn{} paired 95\% intervals of \method{} include
zero and \tkBOursHi{} lie above it. On the 16 English tasks of LongBench v1, the differences of \method{} from BF16 are
\lbvOursQwen, \lbvOursLlama{} and \lbvOursBig{} points on Qwen3-8B, Llama-3.1-8B-Instruct and Qwen3.8-27B, and its
direct differences from RTN$^\star$ are \lbvHtohQwen, \lbvHtoh{} and \lbvHtohBig{} points (Table~\ref{tab:lb1}).
Table~\ref{tab:agentic27} extends the agentic evaluation of Qwen3.8-27B to 16 $\tau^2$-bench trials per task over all
three domains and to all 500 SWE-bench Verified instances, with a second BF16 run as a reference for run-to-run spread.

\subsection{Kernels and Serving}
\label{app:kernels}
The 4-bit cache is a KV-cache data type of vLLM: our cache writer, paged decode-attention kernels and a prefill kernel
run inside vLLM's model runner and are captured in CUDA graphs. Each INT4 group stores its zero point in the four low
mantissa bits of its FP16 scale and its flag in the sign bit, so the page layout is that of symmetric INT4. For keys,
each query head is transformed once per step into every candidate basis, the $qk^\top$ products of all variants are
computed on the tensor cores, and each token selects its products by its stored flags. Prefill chunks with at most 2048
new tokens (4096 for $D{=}256$) attend directly over the INT4 pages of a cached prefix, converting each page tile in
shared memory and applying the per-token inverse transforms there; longer chunks dequantize the prefix to BF16. Every
token is encoded into the pages when it is written, so a full page is a valid prefix-cache block; the BF16 copies of the
$W$ most recent tokens live in a per-request ring and those of the $S$ sink tokens in a pool shared by requests with the
same first tokens. The cache writers are bit-identical to the reference encoder, the decode kernels agree with float64
attention over the dequantized cache within $\max(2\epsilon_{\mathrm{SDPA}}, 4\times10^{-3})$ relative to the largest
output in 410 cases per format, where $\epsilon_{\mathrm{SDPA}}$ is the error of BF16 SDPA on the same inputs, and the
KL of the served cache agrees with the simulated quantizer within 3\% on 20 WikiText-2 windows for the 8B models.

Table~\ref{tab:tput} reports the generation throughput of Qwen3-8B serving $\tau^2$-bench retail agents, multi-turn
tool-use conversations that reuse long prefixes, at 64 to 256 concurrent agents with prefix caching. Against the faster
FP8 backend, \method{}$^\star$ serves at \tpOursFiH$\times$ the throughput on H200 and \tpOursFiL$\times$ on H20, and
the fixed mode at \tpStaticFiH$\times$ and \tpStaticFiL$\times$; with the BF16 sink and window tokens, \method{} runs at
\tpOursWzH$\times$ (H200) and \tpOursWzL$\times$ (H20) of its throughput without them, and it retains \tpOursRtn{} of
the throughput of plain INT4 with our kernels. In the capacity-limited runs where the BF16 cache preempts requests, at
128 and 256 agents on H20 and 256 on H200, \method{} generates \tpOursBfLmid--\tpOursBfL$\times$ and \tpOursBfH$\times$
as many tokens per second as BF16. Each client process drives 8 agents, every server uses one GPU with 90\% of its
memory and a 512-sequence limit, and the user simulators run on separate GPUs. Each point is the mean of two runs on
different GPUs or hosts, or the median of three where their half-range exceeded 10\% of the mean; some H20 points at 256
agents are medians of 3--4 matched repeats. Throughput is generated tokens per second over a timed window after warm-up
(vLLM 0.30.0, PyTorch 2.13 with CUDA 13.0, FlashInfer 0.6.18).

\FloatBarrier
\section{Baselines and Reproductions}
\label{app:audit}
Every external baseline was run from official code or kernels where they exist and checked against a reference
(Table~\ref{tab:audit}); the comparisons of \S\ref{sec:fid} and Appendix~\ref{app:four} rest on these implementations.

\begin{table}[!htbp]
\centering
\small
\setlength{\tabcolsep}{3pt}
\begin{tabular}{@{}lll@{}}
\toprule
Baseline & Source & Check \\
\midrule
vLLM INT4 & vLLM kernels & bit-exact \\
TurboQuant & vLLM kernels & bit-exact \\
Four Over Six & vLLM CUDA code & transcribed \\
KIVI & official code & bit-exact \\
SmoothAttention & QServe code & same calibration \\
QuaRot, SpinQuant & official code & PPL within 1.1\% \\
RotateKV & official code & PPL 5.509/5.511 \\
WUSH-KV & paper & published PPL \\
\bottomrule
\end{tabular}
\caption{How each external baseline was obtained and checked.}
\label{tab:audit}
\end{table}

\paragraph{4-bit baselines.} vLLM's INT4 cache and TurboQuant reproduce vLLM's kernels bit-exactly, including vLLM's
randomized Hadamard and its rounding to BF16; Four Over Six is transcribed from vLLM's CUDA code. KIVI uses its official
quantization functions, which our implementation matches bit-exactly on captured keys and values, together with its
residual policy (keys are quantized once 128 have accumulated); a group whose values are all equal, which occurs on
Qwen2.5-7B and divides by zero in the official code, is kept exact. SmoothAttention follows QServe's calibration with
key maxima after RoPE and $\alpha{=}0.5$. For QuaRot and SpinQuant, we ran the official code with 4-bit keys and values
and BF16 weights and activations on Llama-3.1-8B and Qwen3-8B (Qwen3 ported to both code bases), including SpinQuant's
rotation learning; with the same quantizer and transforms our framework agrees with the official perplexities within
1.1\% in all 14 configurations, and within 0.21\% in 11 of them. WUSH-KV is implemented from its paper, with the QuEST
constant $\alpha_b$ set to the value that minimizes the mean squared error on a standard Gaussian, as its definition
prescribes; Table~\ref{tab:wushrepro} compares our evaluation with its published perplexities.

\begin{table}[!htbp]
\centering
\small
\setlength{\tabcolsep}{4pt}
\begin{tabular}{@{}lrr@{}}
\toprule
4-bit KV cache & WUSH-KV paper & Our evaluation \\
\midrule
No transform & 11.36 & 11.34 \\
Hadamard & 9.77 & 9.77 \\
WUSH & 9.73 & 9.74 \\
\bottomrule
\end{tabular}
\caption{WikiText-2 perplexity of the WUSH-KV evaluation on Qwen3-8B (QuEST, one scale per token and head, 16 sink
and 128 recent tokens, FineWeb-Edu calibration); BF16: 9.72 in the paper, 9.73 here.}
\label{tab:wushrepro}
\end{table}

\paragraph{RotateKV at 2 bits.} With its official settings (grouped-head Hadamard before RoPE, channel reordering,
$g{=}128$, clip ratio 0.8, 20 sink tokens), our RotateKV simulation gives a WikiText-2 perplexity of 5.509 for
Llama-2-7B at 2 bits and 4096 tokens, against 5.511 from the official code. In the protocol-matched runs we keep the
reordering, calibrate it on our four training windows and replace its sink tokens with S4W128.

\FloatBarrier
\section{Mechanism}
\label{app:mech}
Whether a Hadamard rotation lowers the error of a 4-bit key depends on the vector, the model and the format, and a few
stored bits per token capture most of what a per-token selection can gain. This appendix supports the analysis of
\S\ref{sec:background} and the design of Appendix~\ref{sec:method4}.

\subsection{When Rotation Helps}
\label{sec:analysis}
A Hadamard transform spreads a spike over all coordinates, but it can also concentrate the energy of a flat vector. On
INT4 caches of Llama-3.1-8B and Qwen3-8B, $\had_{32}$ raises the query-weighted error of \motFlatUp\% of the flattest
tenth of the key sub-vectors and of \motSpikyUp\% of the spikiest tenth; on NVFP4, with $\had_{16}$ per scale group, the
share is \motNvUp\% across the deciles of the crest factor (Figure~\ref{fig:motivation}b), so the crest factor alone does
not predict the outcome.

\paragraph{Models.} Rotating whole heads, which helps Llama-3.1-8B on the three 4-bit formats, raises the KL of MXFP4 and
NVFP4 caches on Qwen3-8B by \kvMxQwenFull\% and \kvNvQwenFull\% (Table~\ref{tab:kvmainfull}). The harm lies in the keys
and in the E2M1 grid rather than in the E4M3 scales (Table~\ref{tab:nvmech}): with exact FP32 scales the increase on
Qwen3-8B persists (\mechFpQwen\%), whereas a uniform 4-bit grid of the same range turns it into a reduction
(\mechUnifQwen\%). Rotation flattens the key groups, which E2M1 represents worse than spiky ones: its error relative to
the uniform grid is \mechRatioBefore{} before rotation and \mechRatioAfter{} after it. Across the \mechN{} models, the
change of the attention-weighted logit error agrees in sign with the KL change in \mechSign{} cases (Spearman
$\rho=\mechRho$), whereas simple key statistics do not separate the models on which whole-head rotation helps from those
on which it hurts (Table~\ref{tab:keystats}).

\begin{table}[!htbp]
\centering
\fontsize{8}{9.5}\selectfont
\setlength{\tabcolsep}{3.4pt}
\begin{tabular}{lrrrrrr}
\toprule
Quantized & K+V & K & V & K & K & K \\
Group scale & E4M3 & E4M3 & E4M3 & FP32 & E4M3 & FP32 \\
Grid & E2M1 & E2M1 & E2M1 & E2M1 & unif. & unif. \\
\midrule
\multicolumn{7}{l}{\emph{With QK-norm}} \\
Qwen3-8B & \textcolor{gray}{+210} & \textcolor{gray}{+228} & $-$3 & \textcolor{gray}{+264} & \textbf{$-$32} & \textbf{$-$27} \\
Qwen3-14B & \textcolor{gray}{+31} & \textcolor{gray}{+34} & \textcolor{gray}{+4} & \textcolor{gray}{+46} & \textbf{$-$48} & \textbf{$-$44} \\
Gemma-3-27B & \textcolor{gray}{+82} & \textcolor{gray}{+93} & +1 & \textcolor{gray}{+97} & $-$1 & \textbf{$-$3} \\
\midrule
\multicolumn{7}{l}{\emph{Without QK-norm}} \\
Llama-3.1-8B & \textbf{$-$30} & \textbf{$-$35} & \textcolor{gray}{+8} & \textbf{$-$28} & \textbf{$-$44} & \textbf{$-$39} \\
Mistral-7B-v0.3 & \textbf{$-$36} & \textbf{$-$39} & +2 & \textbf{$-$34} & \textbf{$-$43} & \textbf{$-$41} \\
Phi-4 & \textbf{$-$30} & \textbf{$-$33} & 0 & \textbf{$-$30} & \textbf{$-$51} & \textbf{$-$50} \\
\bottomrule
\end{tabular}
\caption{Source of the NVFP4 sign flip: relative KL change (\%) of whole-head Hadamard rotation without BF16 tokens; the unquantized cache stays in BF16 (groups of 16; \emph{unif.}: uniform 15-level grid). Bold / gray: significantly below / above zero (95\% CI).}
\label{tab:nvmech}
\end{table}

\begin{table}[!htbp]
\centering
\fontsize{8}{9.5}\selectfont
\setlength{\tabcolsep}{3pt}
\begin{tabular}{lcrrrrr}
\toprule
 & QK- & ch. & & & \multicolumn{2}{c}{Head rot.\ (\%)} \\
\cmidrule(lr){6-7}
Model & norm & ratio & crest & kurt. & MXFP4 & NVFP4 \\
\midrule
Llama-3.1-8B & no & 7.1 & 4.91 & 8.5 & $-$71 & $-$30 \\
Qwen3-8B & yes & 8.2 & 4.98 & 9.9 & +41 & +215 \\
Mistral-7B-v0.3 & no & 7.1 & 4.72 & 7.8 & $-$62 & $-$35 \\
Qwen3-14B & yes & 7.6 & 4.81 & 9.0 & $-$56 & +32 \\
Qwen3-32B & yes & 6.5 & 4.62 & 8.0 & $-$52 & +26 \\
Qwen3-30B-A3B & yes & 11.0 & 6.28 & 17.0 & $-$19 & +57 \\
Llama-2-7B & no & 9.9 & 6.21 & 15.8 & $-$80 & $-$44 \\
Llama-3.2-3B & no & 7.6 & 5.24 & 9.2 & $-$61 & $-$26 \\
Qwen2.5-7B & no & 8.9 & 5.28 & 10.4 & +4 & +32 \\
Gemma-2-9B & no & 9.1 & 6.27 & 11.6 & $-$70 & $-$38 \\
Phi-4 & no & 10.2 & 6.34 & 15.9 & $-$53 & $-$30 \\
Mistral-NeMo-12B & no & 8.6 & 5.86 & 12.6 & $-$69 & $-$26 \\
GLM-4-9B & no & 10.0 & 5.72 & 15.7 & $-$84 & $-$8 \\
\bottomrule
\end{tabular}
\caption{Post-RoPE key statistics on two validation windows (median over layers and KV heads) and the relative KL change of whole-head rotation. \emph{ch.\ ratio}: largest/median channel RMS; \emph{crest}: max/RMS of a key vector; \emph{kurt.}: its kurtosis over channels.}
\label{tab:keystats}
\end{table}

\begin{figure*}[!tp]
\centering
\figpdf{\textwidth}{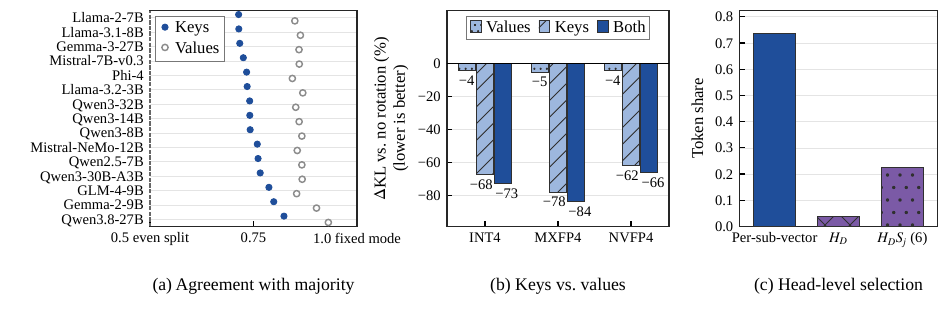}
\caption{(a)~Agreement of the per-token decision whether a sub-vector takes a block transform with its majority
decision (mean over layers, heads and sub-vectors; \mlarge{}, INT4, test windows; 0.5: even split). (b)~Per-token
transforms on values, keys or both (relative KL change w.r.t.\ no rotation; \mlarge{}, validation windows of the
development models). (c)~Share of tokens per head-level selection for keys.}
\label{fig:mech}
\end{figure*}

\paragraph{Tokens.} The best transform also differs between the sub-vectors of one key and from token to token
(Figure~\ref{fig:motivation}c): with a per-token selection among $I$ and $\had_{32}$ per sub-vector and $\had_{128}$
per head, \motShareHead\% of the key sub-vectors of INT4 caches take the whole-head transform, \motShareBlock\% take
$\had_{32}$ and \motShareId\% stay untransformed. For a given (layer, head, sub-vector), the decision of \mlarge{} agrees
with its majority for \bitConsistency{} of the tokens on average (Figure~\ref{fig:mech}a). A transform fixed per head or
sub-vector cannot follow this variation: on validation windows, a static selection among these candidates, made layer by
layer with the true KL divergence, lowers the KL of INT4 caches by \dseStaticKl\%, and a per-token selection among the
same candidates by \dseBlockHead\% (Table~\ref{tab:dse}).

\subsection{Why a Few Stored Bits Suffice}
\label{sec:mech}
Keeping the best of $K$ quantizations of a key costs $\lceil\log_2 K\rceil$ bits, and most of the gain comes from the
first candidates. Across sign-flipped whole-head Hadamard transforms on one validation window of Llama-3.1-8B, the
smaller of two query-weighted errors of a key is on average \bokTwo{} of a single one, and the smallest of 64 is
\bokSixtyFour.

\paragraph{Selection criterion.} On test windows of the development models, per-token selection of the default
candidates by plain reconstruction error lowers the KL of INT4 caches by \ctMse\%, as much as a static selection
(\ctStatic\%); weighted by $C_q$, the same selection lowers it by \ctOurs\%, and by \ctOursSw\% at S4W128, where the
plain-error selection gains \ctMseSw\% (Table~\ref{tab:dsecond}). On NVFP4, query weighting raises the reduction from
\ctMseNv\% to \ctOursNv\% without protected tokens and from \ctMseNvSw\% to \ctOursNvSw\% at S4W128.

\begin{table}[!htbp]
\centering
\small
\setlength{\tabcolsep}{2.9pt}
\begin{tabular}{lrrrr}
\toprule
 & \multicolumn{2}{c}{INT4} & \multicolumn{2}{c}{NVFP4} \\
\cmidrule(lr){2-3}\cmidrule(lr){4-5}
Selection & W0 & S4W128 & W0 & S4W128 \\
\midrule
RTN (KL) & 19.8 & 2.40 & 22.8 & 3.29 \\
\midrule
\multicolumn{5}{l}{\emph{Static}} \\
\quad K only & \textbf{$-$41} & \textbf{$-$37} & \textbf{$-$23} & \textbf{$-$37} \\
\quad K and V (fixed mode) & \textbf{$-$42} & \textbf{$-$39} & \textbf{$-$23} & \textbf{$-$37} \\
\multicolumn{5}{l}{\emph{Per token, K only}} \\
\quad by MSE & \textbf{$-$41} & \textbf{$-$35} & \textbf{$-$23} & \textbf{$-$27} \\
\quad by $C_q$ (\method{}) & \textbf{$-$56} & \textbf{$-$46} & \textbf{$-$46} & \textbf{$-$46} \\
\bottomrule
\end{tabular}
\caption{Transform selection rules on the test windows of the four development models: RTN geometric-mean KL$\times10^3$ and mean KL change (\%) from RTN, with the same candidates in every row. Bold: significantly below RTN on each of the four models (95\% CI).}
\label{tab:dsecond}
\end{table}

\paragraph{Keys, not values.} In the eight-candidate validation ablation, selecting on keys alone recovers \keyShare\%
of the KL reduction of selecting on keys and values, and selecting on values alone lowers KL by \valueOnly\%
(Figure~\ref{fig:mech}b). A key error shifts that token's logit for every later query before the softmax, whereas a
value error enters the output scaled by the token's attention weight. On INT4, keys-only selection lowers KL by
\dseTwoKeys\% with 0.02 stored flag bits per element (Table~\ref{tab:dse}).

\paragraph{Design space.} Table~\ref{tab:dse} covers uniform rotations, static selections per sub-vector or head and
per-token selections with different candidates and criteria. Larger candidate sets add to the gain of \method{} with
diminishing returns (\mlarge{}: \dseCbInt\% at 0.12 bits per element), and each head-level candidate of \mlarge{} is
selected for some tokens, the whole-head transforms for a minority (Figure~\ref{fig:mech}c).

\begin{table}[!htbp]
\centering
\fontsize{8}{9.5}\selectfont
\setlength{\tabcolsep}{2.2pt}
\begin{tabular}{lrrrr}
\toprule
Decision & bits & INT4 & MXFP4 & NVFP4 \\
\midrule
\multicolumn{5}{l}{\emph{Uniform}} \\
Every block & 0.000 & $-$37.0 & $-$51.5 & +4.9 \\
Whole head & 0.000 & $-$35.1 & $-$39.1 & +42.4 \\
\midrule
\multicolumn{5}{l}{\emph{Static (one selection per vector or head)}} \\
Block bits & 0.000 & $-$37.4 & $-$54.5 & $-$8.6 \\
Block bits $|$ whole head & 0.000 & $-$42.2 & $-$59.2 & $-$23.0 \\
Same, by true KL per layer & 0.000 & $-$42.9 & $-$60.6 & $-$23.5 \\
\midrule
\multicolumn{5}{l}{\emph{Dynamic (chosen per token at cache write)}} \\
Block bits, MSE & 0.031 & $-$40.8 & $-$58.9 & $-$22.4 \\
Block bits, $C_q$ & 0.031 & $-$51.4 & $-$68.3 & $-$38.9 \\
Block bits $|$ whole head, $C_q$ & 0.039 & $-$58.9 & $-$74.0 & $-$47.7 \\
Same, attention-weighted & 0.039 & $-$59.6 & $-$75.2 & $-$49.1 \\
Same, $C_q$, keys only (\method{}) & 0.020 & $-$55.9 & $-$69.5 & $-$46.0 \\
Codebook $3{\times}4$, $C_q$ & 0.078 & $-$68.0 & $-$80.3 & $-$60.3 \\
Same, attention-weighted & 0.078 & $-$68.3 & $-$80.9 & $-$60.7 \\
Codebook $8{\times}8$, $C_q$ (\mlarge{}) & 0.117 & $-$72.7 & $-$83.9 & $-$66.2 \\
\bottomrule
\end{tabular}
\caption{Design space of the rotation decision (four development models, 40 WikiText-2 \emph{validation} windows): mean KL change (\%) from no rotation. \emph{bits}: extra bits per element to store the selection ($r_b{=}32$, $D{=}128$).}
\label{tab:dse}
\end{table}

\FloatBarrier

\end{document}